\documentclass[journal]{IEEEtran}
\usepackage[T1]{fontenc}
\usepackage[utf8]{inputenc}
\usepackage{lmodern}
\usepackage{graphicx}
\usepackage{booktabs}
\usepackage{longtable}
\usepackage{array}
\usepackage{calc}
\usepackage{ragged2e}
\usepackage{microtype}
\usepackage{amsmath,amssymb}
\usepackage{xurl}
\usepackage[hidelinks]{hyperref}

\usepackage{tabularx}
\newcolumntype{Y}{>{\RaggedRight\arraybackslash}X}

\renewcommand{\arraystretch}{1.08}

\title{Large Multimodal Agents for Intelligent Transportation Systems:\\
Architectures, Evidence, and Deployment Challenges}

\author{Muhammad Ayub Sabir\textsuperscript{\textdagger},
Shaohong Zheng\textsuperscript{\textdagger},
Zhiyu Qu\textsuperscript{\textdagger},
Fatima Ashraf\textsuperscript{\textdagger},
and Junbiao Pang%
\thanks{\textsuperscript{\textdagger}Muhammad Ayub Sabir, Shaohong Zheng, Zhiyu Qu, and Fatima Ashraf contributed equally to this work.}%
\thanks{Corresponding author: Junbiao Pang (e-mail: junbiao\_pang@bjut.edu.cn).}}

\begin{document}
\maketitle

\begin{abstract}
Large multimodal agents (LMAs) are increasingly proposed for
intelligent transportation systems (ITS), but existing studies often
conflate multimodality, agency, empirical performance, and deployment
readiness. This review provides an auditable evidence map of 42 primary study families released between January 2023 and 3 August 2026 within a corpus of 91 mapped sources. It distinguishes model-level, system-level, and hybrid multimodality and classifies each family by system architecture and action authority. Evidence is assessed independently through functional capability (C0--C3), validation setting (E0--E4), three evidence propositions (P1--P3), and eight methodological-concern domains (Q1--Q8). Transportation semantics (P1) are directly evaluated in 23 families and multidimensional integration (P3) in 24; 19 families directly evaluate both, and all direct P1 or P3 judgments use strong or moderate claim-matched comparisons. Evidence reconciliation (P2) remains unresolved because no family demonstrates the complete provenance--challenge--handling--comparison--outcome chain. Capability also exceeds validation maturity. Fourteen families reach C3, but 13 remain at E2; only one reaches E3, none reaches E4, and no family provides substantive robustness and failure evidence under Q6. Across ITS domains, LMAs are best supported for semantic interpretation, intent translation, evidence organisation, scenario authoring, explanation, and specialist-tool coordination. Numerical forecasting, optimisation, simulation fidelity, hard constraints, low-level control, safety fallback, and final authority should remain with independently verifiable specialist systems or accountable humans. The review therefore supports bounded orchestration rather than replacement and provides a matched comparative evaluation protocol and staged roadmap for accountable deployment. The living evidence repository is available at \url{https://github.com/pangjunbiao/ITS-LMA-Review}.
\end{abstract}


\section{Introduction}\label{introduction}

\subsection{Motivation, Review Gap, and Research Questions}
\label{motivation-and-research-questions}

Systems based on foundation models may complement established transportation models by interpreting heterogeneous evidence and coordinating specialist tools. Some systems process multiple modalities within one model; others achieve system-level multimodality by retrieving or combining evidence from maps, databases, sensor modules, simulators, forecasters, planners, optimisers, or controllers. Their functional roles also differ. A system may provide offline
perception or grounding, support a human decision, execute a bounded
action, or use an observed transportation outcome to revise a later
decision. Comparing all such systems under a single ``agent'' label
would therefore obscure important differences in architecture,
authority, feedback, and empirical maturity.

Recent surveys cover general LMA architectures, multimodal and
vision--language--action systems for autonomous driving, and broader
applications of large language models in transportation
\cite{ref4,ref6,ref9,ref13,ref58,ref61}. They discuss multimodal
reasoning, agent components, technical performance, safety, governance,
and deployment challenges. As examined in
Section~\ref{related-surveys-and-unresolved-review-gap}, however, the
reviewed survey literature does not jointly operationalise four
distinctions needed for an evidence-based comparison: model-level
versus tool-mediated system-level multimodality; functional capability
versus validation setting; proposition directness versus favourable
result direction; and LMA functionality versus the authority retained
by specialist tools, human operators, and independent safety
mechanisms. This unresolved analytical gap motivates the present
review.

The review focuses on foundation-model-centred systems addressing a
defined ITS task and released from January 2023 through 3 August 2026,
while retaining earlier sources needed to define terminology,
specialist baselines, datasets, simulators, and assurance requirements.
The primary unit of analysis is a verified study family rather than an
individual paper version. The final evidence map contains 42 primary
study families within 91 mapped sources and examines technical system
form, capability, validation, proposition evidence, methodological
concern, evaluation resources, and operational authority. The detailed
review process is reported in Section~\ref{review-methodology}.

The central question is not whether an LMA can replace established
transportation models, but where its semantic interpretation and tool-
orchestration capabilities provide measurable value under a
claim-matched comparison. Depending on the proposition, the comparison
may be an appropriately matched specialist baseline, a component
ablation, a controlled semantic perturbation, or a reduced-input or
nonintegrated condition. Such value must be interpreted together with
provenance, uncertainty, robustness, latency, action authority, human
oversight, and operational safeguards. Accordingly, the review asks:
\emph{RQ1}, in which ITS tasks do transportation semantics and
multidimensional evidence integration provide directly or partially
supported value under claim-matched comparison, and where does evidence
remain insufficient? \emph{RQ2}, which ITS-LMA system forms, functional
capabilities, action-authority boundaries, validation settings, and
technical evidence patterns have been demonstrated? \emph{RQ3}, how
should responsibility be divided among the LMA, specialist forecasting,
optimisation, simulation, planning, and control tools, human operators,
and independent safety mechanisms? \emph{RQ4}, which evidence gaps and
technical, operational, safety, reproducibility, and governance gates
must be closed before greater action authority can be justified?

\subsection{Contributions}
\label{contributions}

This review makes six contributions.
\begin{enumerate}
\item It establishes a broad but operational definition and inclusion
boundary for ITS-LMAs. The definition distinguishes model-level,
system-level, and hybrid multimodality; identifies the substantive
foundation-model role; and separates eligible systems from language
interfaces, conventional multimodal predictors, and contextual
resources.

\item It provides a reproducible structured-review protocol based on
study-family resolution, cutoff-valid source/version verification,
source-linked evidence locators, pilot calibration, label-masked repeat
audits, and a frozen canonical synthesis database. This protocol links
42 primary families to 68 primary-family source/version records without
double counting overlapping reports.

\item It develops an orthogonal study-level evidence framework. Three
testable propositions---behaviour and facility semantics (P1), evidence
reconciliation (P2), and multidimensional transportation integration
(P3)---are assessed separately from the C0--C3 functional capability
scale, the E0--E4 validation-setting scale, the noncompensatory Q1--Q8
methodological-concern profile, and result direction. Supporting
technical descriptors cover model type, system architecture,
multimodality, tools, memory, authority, feedback, adaptation, and human
involvement.

\item It constructs a verified technical and cross-domain evidence map
covering exact modality combinations, system families, datasets,
simulators, primary evaluation environments, comparator strength,
robustness, and code, data, and prompt or configuration availability.
The synthesis identifies 19 families with direct P1 and P3 evidence,
nine with one direct proposition, and 14 with partial evidence for both,
without converting these positions into a quality score or study
ranking.

\item It derives an evidence-based allocation of responsibility among
LMAs, specialist forecasting, optimisation, simulation, planning, and
control systems, human operators, and independent safety mechanisms.
The resulting operational architecture preserves explicit boundaries
for provenance, uncertainty, permissions, approval, monitoring,
feedback, fallback, and authorised action.

\item It proposes a matched comparative evaluation and ablation protocol
and a staged deployment roadmap. The protocol links specialist
baselines, semantic and integration ablations, operational envelopes,
action authority, robustness, safety assurance, fallback, rollback,
cybersecurity, audit, and post-deployment monitoring to the evidence
required for progressively greater system responsibility.
\end{enumerate}

These contributions also have direct stakeholder implications.
For traffic operators, road users, public agencies, and service providers,
potential gains in evidence synthesis, context-aware support, planning, and
tool coordination must be balanced against automation bias, unequal sensing
coverage, privacy and cybersecurity risks, vendor dependence, and unclear
accountability through appropriate human oversight, provenance, audit,
fallback, and incident-response mechanisms.

\subsection{Paper Organisation}
\label{paper-organisation}

Section~\ref{review-methodology} details the review design and coding,
while Section~\ref{conceptual-framework-of-lmas-in-its} defines the
ITS-LMA boundary, propositions, capability and validation scales, and
operational architecture. Section~\ref{related-surveys-and-unresolved-review-gap}
positions the review against prior surveys, and
Section~\ref{evidence-across-intelligent-transportation-scenarios}
synthesises the five application domains.
Section~\ref{cross-scenario-limitations-resources-and-deployment-roadmap}
addresses assurance, reproducibility, the research questions, and the
deployment roadmap. The final sections present limitations and
conclusions; detailed study-level and audit records are provided in the
supplementary material.

\section{Review Methodology}
\label{review-methodology}
\label{scope-and-structured-review-protocol}

\subsection{Review Design and Corpus Construction}
\label{review-design-and-corpus-construction}

This study uses a structured narrative review and evidence-mapping
design rather than claiming an exhaustive systematic review. The main
search window covers work published or publicly released from January
2023 through 3 August 2026. Earlier publications were retained only when
needed to define terminology, specialist baselines, datasets,
simulators, standards, or deployment requirements. The study family,
not the individual paper, is the primary analytical unit: preprints,
conference papers, and journal extensions describing the same core
system and substantially overlapping evaluation were linked and counted
once.

The original search covered IEEE Xplore, the ACM Digital Library,
ScienceDirect and DOI-linked publisher pages, SpringerLink, Taylor \&
Francis Online, the CVF Open Access repository, and supplementary arXiv
records. A targeted repeat search was completed on 3 August 2026 using
topic queries, named-system and exact-title checks, backward and forward
citation tracing, and publication-version verification across
DOI-linked publisher pages, CVF Open Access, OpenReview, PMLR, and arXiv.
The common search concept was

\begin{quote}
(``large language model'' OR ``multimodal large language model'' OR
``vision--language--action'' OR ``foundation model'' OR agent*)
AND
(traffic OR transportation OR ``autonomous driving'' OR
``signal control'' OR safety OR planning OR simulation).
\end{quote}

Scenario-specific terms were added for traffic prediction, route
choice, public transport, incident analysis, safety, traffic control,
planning, and transportation simulation. Backward and forward citation
checking was used to identify records not retrieved by the common query.

A record was eligible as primary ITS-LMA evidence when it: (1) contained
a substantive foundation-model component; (2) addressed a defined
transportation problem; (3) integrated heterogeneous transportation
evidence at the model level, system level, or both, according to the
operational boundary in Section~\ref{sec:operational_definition}; and
(4) used the foundation-model component for transportation-specific
grounding, reasoning, retrieval, tool orchestration, decision support,
planning, simulation, or authorised action. C0 systems were retained
only when they demonstrated transportation-specific multimodal
foundation-model capabilities that provide a substantive enabling
function for later decision support or agency.

The primary map excludes interfaces that merely verbalise numerical
outputs, conventional multimodal predictors without a substantive
foundation-model component, generic demonstrations without a defined
transportation role, duplicate versions, and reports lacking sufficient
methodological evidence. Conventional models, datasets, simulators,
standards, governance documents, and safety-validation studies were
retained only as contextual baselines, resources, or deployment anchors.
Closely overlapping reports with the same architecture, capability, and
validation pattern were consolidated into one family; materially
different architectures or action cycles were retained separately.
Thus, DriveGPT4 and DriveGPT4-V2, and ChatSUMO and ChatSUMO-Agent, were
retained as distinct families, whereas linked GATSim, AgentSUMO/Speak to
Simulate, and Virtual Traffic Police reports were consolidated within
their respective families.

Every provisionally eligible family underwent source-level full-text
verification by one trained reviewer. Eligibility, family membership,
publication status, multimodality, foundation-model role, executor and
authority boundary, feedback path, capability, validation, proposition
evidence, methodological concern, and the narrowest available source
locator were rechecked. Ambiguous cases received the lower demonstrated
classification. The final closure screen added one controlled
real-world autonomous-driving family and one highway
incident-management family and retained ProSim only as a contextual
specialist simulator. No eligibility or family decision remained
unresolved when the evidence map was frozen on 3 August 2026.

The final map contains 91 sources: 42 primary study families, 15 related
surveys, 18 contextual baselines or concepts, eight governance or standards
sources, six datasets, benchmarks, or simulators, and two review or
terminology sources. The 42 primary families are linked to 68 cutoff-valid
primary-family source/version records. Detailed corpus composition and
source-role accounting are provided in Supplementary Table~S8, while
family/version decisions and update rules are documented in Supplementary
Sections S2 and S7.

Database-specific retrieval and deduplication totals were not preserved
during the review's earlier iterative development; therefore, a
retrospective PRISMA flow is not reconstructed. The repeat-search
queries, candidate reports, family links, screening decisions, reasons
for exclusion or consolidation, closure decisions, and numerical
reconciliation were instead preserved in the audit register. The
resulting family counts describe the audited analytical evidence map and
should not be interpreted as estimates of publication prevalence.

\subsection{Evidence Coding and Synthesis}
\label{evidence-coding-and-synthesis}

Each primary family was extracted into a common record covering the
transportation task, publication status, eligible system configuration,
model and system architecture, multimodality and input modalities,
tools, memory, action authority, feedback, adaptation, human
involvement, datasets, simulators, evaluation environment, comparators,
quantitative results, reported failures or limitations, and public
artefacts. Capability, validation, proposition evidence, methodological
concern, and result direction were coded independently; none could
compensate for or mechanically determine another.

Capability followed a prespecified C0--C3 decision sequence: substantive
transportation role; participation in an actionable decision-support or
action cycle; execution or governance of a bounded transportation
action; receipt of transportation-state or outcome feedback at the
foundation-model decision layer; and evidence that such feedback changed
a later transportation decision. Supporting descriptors---model type,
system family, multimodality, tool use, memory, authority, feedback,
adaptation, and human role---characterised architecture but did not form
an additive capability score. A 12-family difficult-case pilot clarified
the manual without changing any C assignment. A label-masked repeat
audit reproduced 40 of 42 capability labels before adjudication; the two
boundary disagreements were resolved using the frozen rules.

Validation was coded independently on the E0--E4 scale, distinguishing
no task-level empirical evaluation, fixed or offline evidence,
responsive simulation or virtual operation, controlled real-world
evaluation, and sustained routine deployment. Real-world recorded data,
physical hardware, simulator fidelity, or the term ``deployment'' did
not substitute for the required environment, attribution, protocol, or
duration evidence. Methodological concern was assessed independently
across eight noncompensatory domains: task/data appropriateness (Q1),
implementation transparency (Q2), comparators and metrics (Q3),
analysis and uncertainty (Q4), external validity (Q5), robustness and
failure analysis (Q6), reproducibility and traceability (Q7), and
authority, safety, and governance (Q8). Calibration used 12 difficult
families and four boundary vignettes. The label-masked repeat audit
reproduced 37 of 42 E labels and 280 of 336 Q judgments before
adjudication; no final E label changed, while three Q judgments changed
from high to some concern. Q judgments were neither aggregated nor used
to rank studies.

P1--P3 evidence was coded independently of capability, validation,
methodological concern, publication status, and system multimodality.
\emph{D} denotes direct evaluation of a proposition, \emph{P}
substantive but incomplete or indirect evaluation, and \emph{N} no
proposition-specific evaluation. Result direction was recorded
separately as positive, mixed, null, negative, or not reported; thus,
D does not imply benefit and N does not imply failure. P1-D required an
explicit transportation-semantic element, its functional use, a
claim-matched comparison or perturbation, an attributable
transportation outcome, and contribution attribution. P2-D required
traceable provenance, an evaluated missingness, uncertainty, or
source-conflict challenge, a handling mechanism, a matched comparison,
and an attributable outcome. P3-D required at least two substantively
distinct transportation evidence dimensions, their actual integration,
a matched reduced-input, nonintegrated, misaligned, factorial, or
component comparison, an attributable outcome, and integration
attribution.

The proposition construct and evidence manuals were calibrated on 12
difficult families and six boundary vignettes before all 126
family--proposition judgments were recoded. The label-masked repeat
audit reproduced 113 of 126 judgments before adjudication, with three
initial D judgments revised to P. The final register contains 66
high-confidence and 60 moderate-confidence judgments, with no
low-confidence or unresolved final code. Controlled experiments and
structured comparative human or operational studies could support D;
implementation evidence could support P; abstracts, demonstrations,
author claims, and secondary summaries could not determine a final code
alone.

After coding was frozen, a canonical one-row-per-family synthesis
database was constructed, with the audited proposition register serving
as the analytical source of truth rather than manuscript prose. Each
family was linked to its eligible configuration, cutoff-valid sources,
technical descriptors, C/E/P/Q profile, result direction, evidence
locators, and confidence judgments. Exact modalities, named data
sources, simulators and operational environments, primary evaluation
environment, comparative strength, and availability of code, data, and
prompts or configurations were verified from the principal paper,
official project page, or official repository. ``Not located'' means
that no official public artifact was found in the searched sources; it
does not imply that no private or later artifact exists.

For cross-domain synthesis, each family was assigned to one of five
mutually exclusive application groups according to its principal
evaluated LMA function rather than title alone. Counts and
cross-tabulations were generated from the canonical database, and
comparator strength was retained as strong, moderate, or
partial/nonisolating under the frozen claim-matched-comparison rules.
Robustness, reproducibility, and governance summaries interpret Q6, Q7,
and Q8 rather than introduce new scores. Because task, dataset, and
metric heterogeneity precluded meta-analysis, the review uses structured
comparative synthesis of direct, partial, and absent evidence without a
composite study ranking.

All pilot and repeat procedures were within-process stability checks,
not independent duplicate human coding; no inter-rater reliability
statistic is claimed. Structured model assistance supported literature
retrieval, metadata checking, extraction drafting, and consistency
auditing, while final eligibility, coding, adjudication, and
interpretation remained under authorial responsibility.

\section{ITS-LMA Framework and Technical Foundations}
\label{conceptual-framework-of-lmas-in-its}

This section establishes the conceptual and technical basis used throughout the
review. It first explains why transportation reasoning may require more than a
fixed feature representation, then defines the ITS-LMA system boundary and
system-family taxonomy, separates functional capability from validation
setting, and finally specifies the technical and operational boundaries of a
deployment-oriented architecture.

\subsection{Evidence-Grounded Transportation Reasoning}
\label{evidence-grounded-transportation-reasoning}

A conventional transportation model maps selected traffic states,
network structure, and exogenous variables to a prediction,
recommendation, plan, or control output. This abstraction is efficient
when tasks, observations, constraints, and operating conditions are
well defined. Its limitation is not numerical processing, but
representing operational meaning distributed across heterogeneous and
partly unstructured evidence. For example, a lane closure, temporary
access restriction, pedestrian-generating event, inaccessible transit
stop, or disagreement among camera, probe-vehicle, map, and operator
records may change the interpretation of an otherwise similar numerical
state.

An ITS-LMA may complement specialist models by constructing an
\emph{evidence-grounded semantic state}: a representation linking
transportation entities, relations, rules, temporal claims, provenance,
uncertainty, and unresolved conflict. Systems may realise this
abstraction through prompts, structured memory, retrieval records,
graphs, tool outputs, or intermediate representations. Such a state
should be treated as a decision-support hypothesis rather than verified
truth, causal explanation, or automatic repair of missing or
conflicting evidence.

The review evaluates three candidate benefits of this semantic layer.
\emph{P1---behaviour and facility semantics} asks whether
transportation meaning---including actors, interactions, rules, intent,
behavioural state, facility function, and event context---materially
changes an inference, recommendation, plan, or action.
\emph{P2---evidence reconciliation} asks whether traceable provenance
and the handling of missingness, uncertainty, or source conflict affect
a transportation task or decision.
\emph{P3---multidimensional transportation integration} asks whether
integrating substantively distinct spatial or network, temporal,
operational, environmental, behavioural, facility, perceptual, or
textual evidence contributes to a defined task. Thus, P1 concerns the
effect of transportation meaning, whereas P3 concerns the effect of
integrating distinct evidence dimensions. TRIP similarly motivates
progressive transportation reasoning \cite{ref3}; here, these benefits
are treated as testable propositions rather than inherent properties of
foundation models.

Multimodality, model scale, tool use, simulation, or whole-system
superiority alone does not establish any proposition. Direct, partial,
and unevaluated evidence follow the claim-matched comparison and
attribution rules defined in the review methodology, while result
direction is recorded independently. The propositions therefore
structure the review's scientific questions without presuming that
semantic integration benefits every transportation task.

\subsection{Operational Definition and System Taxonomy}
\label{sec:operational_definition}

The terms \emph{foundation model}, \emph{multimodal foundation model},
\emph{language agent}, \emph{large multimodal agent}, and
\emph{vision--language--action system} are related but not
interchangeable. A foundation model may be neither multimodal nor
agentic; a multimodal foundation model directly processes multiple data
modalities; a language agent may reason, retrieve information, maintain
context, or call tools while remaining text centred; and a
vision--language--action system connects multimodal perception and
language-conditioned reasoning to an action pathway. These systems
therefore differ in where multimodal integration occurs, how actions are
generated, where implementation authority resides, and how their claims
should be validated.

This review defines multimodality at the level of the evaluated
foundation-model-centred system. \emph{Model-level multimodality} means
that the central model directly processes at least two transportation
modalities, such as text, images, video, maps, trajectories, traffic
time series, LiDAR, or radar. \emph{System-level multimodality} arises
when a language-centred model accesses and integrates heterogeneous
evidence through databases, maps, sensor-processing modules, retrieval
systems, simulators, forecasters, optimisers, planners, or controllers.
\emph{Hybrid multimodality} combines both. This distinction includes
text-centred models with substantive multimodal tool access without
misclassifying conventional multimodal predictors that lack a
foundation-model reasoning or decision role.

An ITS-oriented large multimodal agent is therefore defined as a
foundation-model-centred transportation system in which the foundation
model has a substantive role in interpreting heterogeneous evidence,
reasoning under transportation context, coordinating specialist tools,
supporting decisions, generating plans or simulation artefacts,
governing authorised actions, or providing task-specific training,
reward, or supervision. The definition accommodates heterogeneous
terminology in the literature while remaining bounded by the
demonstrated role of the foundation model in the evaluated workflow.

The primary evidence map excludes language interfaces that merely
verbalise specialist outputs, conventional multimodal predictors without
a substantive foundation-model component, generic demonstrations
without a defined transportation role, and contextual resources that do
not evaluate an eligible system. Datasets, simulators, standards,
specialist models, and safety-validation studies are retained only as
contextual evidence unless the foundation-model-centred configuration
itself is evaluated.

Eligible systems are assigned to one principal architectural family
according to the evaluated configuration and foundation-model role:
(i) direct foundation-model inference or reasoning;
(ii) foundation-model integration with a specialist transportation
pipeline;
(iii) a single tool-using foundation-model agent;
(iv) multi-agent foundation-model orchestration; or
(v) foundation-model support for training, reward generation, or
supervision. These families describe system organisation rather than
performance or maturity.

Each family is also characterised independently by model type,
multimodality, tool use, memory, action authority, feedback, adaptation,
and human involvement. Tool use ranges from none to retrieval,
analytical or simulation tools, and action-executing transportation
tools; memory ranges from transient or unreported context to persistent
retrievable or writable experience. Action authority is recorded as no
action authority (A0), external implementation (A1), bounded delegated
execution (A2), or repeated execution within a defined operational
envelope (A3). Feedback distinguishes software or evaluation feedback,
transport-state feedback, and feedback that demonstrably changes a
later foundation-model-level decision. These descriptors are
noncompensatory: greater architectural complexity, tool use, memory,
multi-agent organisation, or multimodal richness cannot by themselves
raise a capability level.

\begin{table*}[!t]
\centering
\caption{Operational capability levels and decision boundaries for
ITS-LMA systems.}
\label{tab:capability-levels}

\footnotesize
\setlength{\tabcolsep}{3pt}
\renewcommand{\arraystretch}{1.12}

\begin{tabular}{@{}
>{\RaggedRight\arraybackslash}p{0.09\textwidth}
>{\RaggedRight\arraybackslash}p{0.21\textwidth}
>{\RaggedRight\arraybackslash}p{0.23\textwidth}
>{\RaggedRight\arraybackslash}p{0.26\textwidth}
>{\RaggedRight\arraybackslash}p{0.15\textwidth}
@{}}
\toprule

\textbf{Level} &
\textbf{Operational definition} &
\textbf{Decisive evidence} &
\textbf{Insufficient evidence} &
\textbf{Typical authority} \\

\midrule

C0: Pre-agentic capability &
Transportation-specific representation, prediction, perception,
grounding, explanation, question answering, offline generation, or
training support without an actionable decision-support or action cycle &
Task-level evidence that the foundation model performs a substantive
transportation-specific function &
An ``agent'' label, multi-step prompting, open-loop control prediction,
or an independently acting downstream controller &
A0: no action authority \\

C1: Decision support &
Actionable retrieval, analysis, planning, tool use, or recommendation
with implementation retained externally &
A defined decision-support output and explicit external implementation
authority &
Architectural complexity or tool use alone; an unexecuted output does
not establish C2 &
A1: external implementation \\

C2: Bounded action &
The system executes, invokes, or governs a bounded virtual or physical
transportation action without demonstrated outcome-driven later
re-decision &
An attributable simulator intervention, policy primitive, controller
parameter, generated programme, or authorised action command &
Software retry, user revision, evaluation scores, or transport-state
observation without evidence of a changed later decision &
A2: bounded delegated execution \\

C3: Outcome-responsive agency &
Observed transportation state or outcomes enter the
foundation-model-centred decision process and change a later
transportation decision &
A traceable sequence from executed action to transport feedback,
updated state, context, or memory, and a changed subsequent decision &
Repeated action, periodic observation, self-reflection, or feedback
confined to a specialist controller without a changed
foundation-model-level decision &
A2--A3: bounded or repeated delegated execution, possibly with human
supervision \\

\bottomrule
\end{tabular}

\end{table*}

\subsection{Capability and Validation Framework}
\label{capability-and-validation-framework}

Functional capability and validation setting answer different questions.
Capability records what the foundation-model-centred system demonstrably
does in the transportation decision pathway, whereas validation records
the environment in which that capability was evaluated. Neither scale
measures predictive performance, methodological quality, safety, or
deployment readiness.

Capability follows a sequential C0--C3 rule. After confirming a
substantive transportation role, C0 applies when the foundation model
does not participate in an actionable decision-support or action cycle.
C1 rule. After confirming a
substantive transportation role, C denotes actionable support with implementation retained by a human or
external system. C2 requires execution, invocation, or governance of a
bounded transportation action. C3 requires observed transportation
state or outcome feedback to reach the foundation-model decision process
and demonstrably change a later transportation decision. Software
retries, self-critique, benchmark scores, user edits, repeated
observations without changed decisions, or feedback confined to an
independently acting specialist controller do not establish C3. Table \ref{tab:capability-levels} summarises the operational definitions and decision boundaries
for the four capability levels.

Validation is coded independently using E0--E4. E0 denotes no task-level
empirical evaluation, E1 fixed or offline evidence, E2 responsive
simulation or virtual operation, E3 controlled real-world evaluation,
and E4 sustained routine deployment. E2 requires a traceable system
output or artefact to cause state transitions in a responsive simulator,
digital twin, emulator, or virtual transportation process. E3 requires
contemporaneous evaluation in a real physical or operational
environment. E4 additionally requires sustained operation within a
declared operational envelope, supported by longitudinal outcomes,
monitoring, intervention or incident handling, and governance evidence. Table \ref{tab:validation-levels} summarises the corresponding validation settings and evidence
boundaries.

\begin{table*}[!t]
\centering
\caption{Validation settings and evidence boundaries for ITS-LMA claims.}
\label{tab:validation-levels}

\footnotesize
\setlength{\tabcolsep}{3pt}
\renewcommand{\arraystretch}{1.12}

\begin{tabular}{@{}
>{\RaggedRight\arraybackslash}p{0.10\textwidth}
>{\RaggedRight\arraybackslash}p{0.23\textwidth}
>{\RaggedRight\arraybackslash}p{0.25\textwidth}
>{\RaggedRight\arraybackslash}p{0.24\textwidth}
>{\RaggedRight\arraybackslash}p{0.12\textwidth}
@{}}
\toprule

\textbf{Level} &
\textbf{Operational setting} &
\textbf{Decisive evidence} &
\textbf{Insufficient evidence} &
\textbf{Interpretation} \\

\midrule

E0: Conceptual &
No task-level empirical evaluation of the eligible ITS-LMA system or
attributable pipeline &
Architecture, use case, qualitative framework, or unmeasured
demonstration only &
Experiments on another component or intended future evaluation &
Conceptual plausibility only \\

E1: Offline evidence &
Fixed datasets, historical records, replay, static benchmarks, offline
generated outputs, or noninteractive artefact evaluation &
A defined task and evaluated configuration with fixed inputs,
outputs, and observed offline results &
Real-world data origin, real-time inference speed, fixed
simulator-generated data, physical hardware, or offline code execution
alone &
Supports the reported offline task only \\

E2: Interactive simulation &
System outputs, policies, parameters, scenarios, or artefacts cause
state transitions in a responsive simulator, digital twin, emulator,
or virtual transportation process &
A traceable output--virtual-response--outcome sequence attributable to
the eligible configuration &
Static simulator data, prerecorded replay, rendering alone, software
retry, or a non-transport code-evaluation loop &
Supports virtual system behaviour; real-world validity remains open \\

E3: Controlled operational evidence &
The integrated eligible system is evaluated contemporaneously in a
real transportation environment under restricted, supervised,
shadow-mode, test-track, or limited-pilot conditions &
A defined real-environment protocol with system-in-the-loop operation
and recorded outputs, outcomes, interventions, or comparisons &
Offline field data, onboard latency tests, physical hardware without
transportation interaction, or anecdotal demonstrations &
Supports bounded operational feasibility, not sustained deployment \\

E4: Sustained deployment evidence &
Routine or repeated real-world operation over a task-appropriate period
within a declared operational envelope and authority boundary &
Documented duration and scope, longitudinal outcomes, continuous
monitoring, incident or intervention handling, governance, and audit
evidence &
A short pilot, one-off demonstration, repeated simulation, or an
unsupported deployment claim &
Supports deployment only within the documented envelope, duration, and
authority \\

\bottomrule
\end{tabular}

\end{table*}

The validation level reflects the highest setting directly demonstrated,
not overall study quality. Retrospective real-world records and fixed
simulator-generated data remain E1; hardware-in-the-loop is normally
E2; live shadow mode may qualify as E3 when contemporaneous outputs and
operational outcomes are systematically recorded; and a limited pilot
is E3, not E4. When the required environment, attribution, protocol, or
duration evidence is absent, the lower level is assigned.

Capability, validation, proposition evidence, result direction, and
methodological quality remain orthogonal. P1--P3 records whether a
specific proposition is directly, partially, or not evaluated; result
direction records whether the observed effect is positive, mixed, null,
or negative; and Q1--Q8 records claim-specific methodological concerns.
A system may therefore demonstrate C3 functionality in E2 simulation
while retaining partial proposition evidence, weak robustness analysis,
or unresolved governance concerns. No dimension compensates for
another, and no composite study ranking is produced.

\subsection{Technical Architecture and Operational Boundaries}
\label{technical-architecture-and-operational-boundaries}

A deployment-oriented ITS-LMA begins with heterogeneous transportation
evidence rather than a single undifferentiated prompt. Text, images,
video, traffic time series, trajectories, maps, LiDAR, radar,
infrastructure records, and operator reports are processed through
modality-specific encoders or specialist modules, then aligned by
spatial, temporal, and map grounding. The resulting semantic state
should retain source identity, observation time, confidence,
missingness, and unresolved disagreement so that generated content
cannot silently replace evidence.

The foundation-model layer performs semantic interpretation, task
decomposition, evidence selection, retrieval, and coordination.
Retrieved rules, historical cases, maps, manuals, operational records,
and remembered context remain evidence to be checked rather than
authority to act. Persistent memory may support later decisions but
introduces risks from stale information, cross-episode contamination,
privacy leakage, and untraceable updates.

Numerical forecasting, optimisation, routing, simulation, motion
planning, risk estimation, and low-level control remain specialist
functions. The LMA may formulate tasks, select tools, construct inputs,
interpret outputs, and compare alternatives, while specialist
components retain explicit objectives, physical constraints,
calibration, and numerical execution. Their outputs should return with
assumptions, validity conditions, uncertainty, and tool versions to
preserve auditability and prevent generated explanations from
substituting for computation.

\begin{figure*}[!t]
\centering
\includegraphics[width=\textwidth]{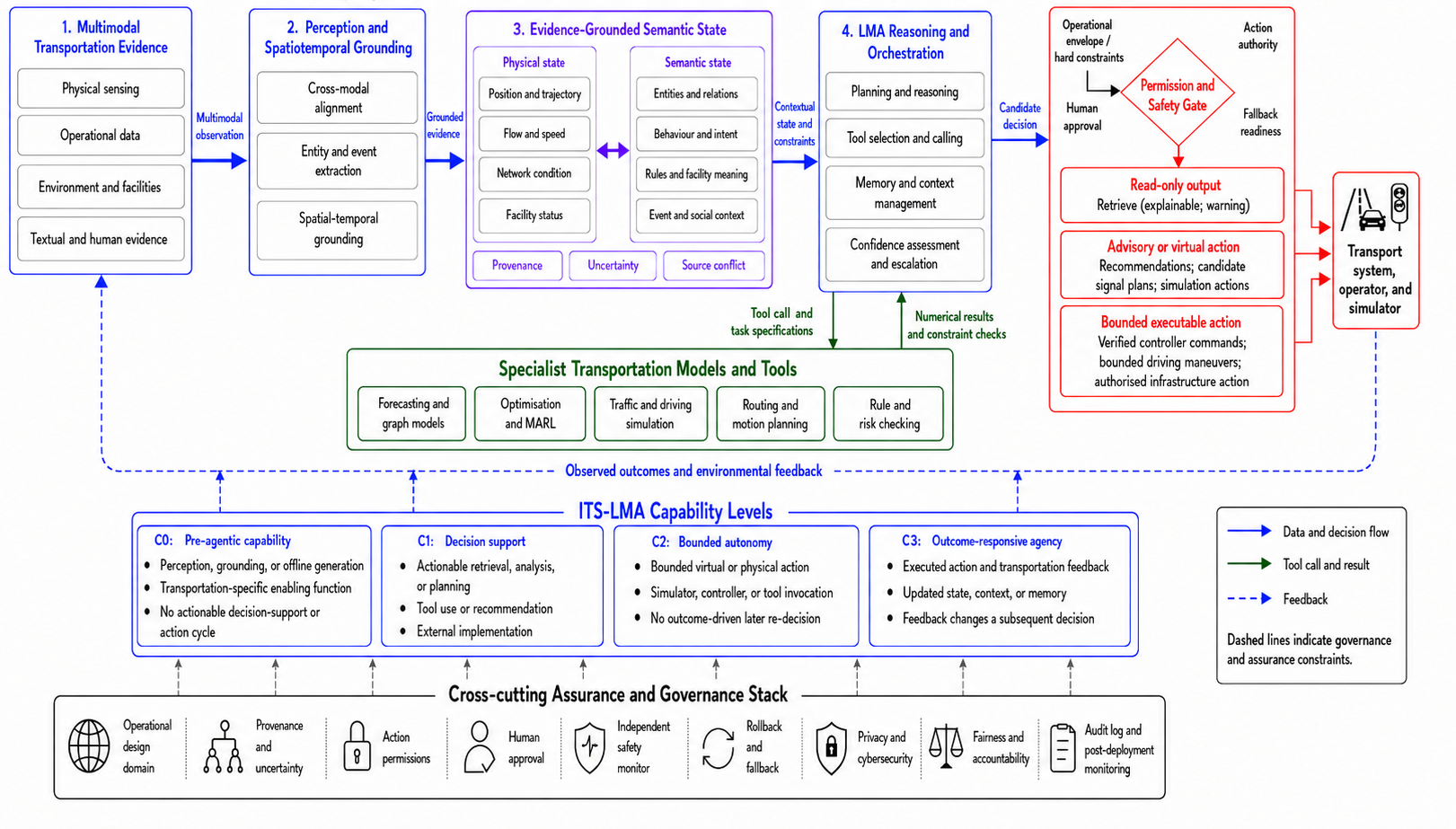}
\caption{Deployment-oriented ITS-LMA reference architecture and operational
boundaries. Multimodal evidence is grounded into a semantic state preserving
provenance, uncertainty, and source conflict. The LMA coordinates specialist
transportation models and tools, while candidate actions pass through
permission and safety checks and observed outcomes inform subsequent
decisions. C0--C3 denotes analytical capability; assurance and governance
apply across capability levels and do not determine capability classification.}
\label{fig:its-lma-architecture}
\end{figure*}

Reasoning outputs are advisory or candidate actions until they pass a
permission and safety gateway that checks authority, operational
envelope, latency, data validity, and hard transportation constraints.
Where errors may have material consequences, human approval and an
independent safety monitor should remain outside the LMA. The monitor
must be able to reject actions, invoke verified fallback, and record
events independently of the model that generated the action.

Outcome-responsive agency begins only after an authorized action or
external decision affects the transportation environment. Observed
outcomes must be distinguished from predicted or self-reported success,
linked to the executed action, and incorporated into state, context, or
memory so that they can change a later decision. This requirement
separates C2 bounded execution from C3 outcome-responsive agency and
prevents software retries, user edits, or simulator errors from being
misclassified as transportation feedback.

Operational authority must remain within an operational design domain
for vehicle automation or, more generally, an \emph{operational
envelope} for other ITS applications. The envelope should specify
geographic and network scope, traffic and weather conditions, sensing
and data-latency requirements, permitted actions, model and tool
versions, human-supervision conditions, fallback resources, and
excluded cases. This follows the bounded-domain principle of SAE J3016
without redefining its vehicle-specific terminology \cite{ref2}.
Deployment design must also address end-to-end latency, communication
failure, model and tool availability, token and energy cost,
edge--cloud dependency, and graceful degradation. Safety-critical
fast loops should default to verified specialist controllers whenever
foundation-model latency or uncertainty cannot be bounded.

Fig.~\ref{fig:its-lma-architecture} summarises this deployment-oriented
reference architecture. It is an author-proposed synthesis of the
evidence and assurance requirements identified in this review and does
not imply that every reviewed system implements every component.

The framework therefore does not assume that semantic models outperform
specialist transportation models, that generated explanations establish
causality, that missing evidence can be freely synthesized, or that
simulation demonstrates deployment readiness. Under current evidence,
LMAs are best supported for interpreting context, retrieving and
organizing evidence, coordinating specialist tools, and supporting
accountable decisions. Numerical prediction, optimization, simulation
fidelity, hard constraints, low-level control, safety fallback, and
final operational or legal authority should remain with specialist
systems or accountable humans unless stronger claim-matched and
operational evidence justifies greater LMA authority.

\begin{table*}[!t]
\centering
\caption{Comparison of related-survey families and the present review.}
\label{tab:related-surveys}

{\footnotesize
\setlength{\tabcolsep}{2.5pt}
\renewcommand{\arraystretch}{1.10}

\begin{tabular}{@{}
>{\RaggedRight\arraybackslash}p{0.14\textwidth}
>{\RaggedRight\arraybackslash}p{0.15\textwidth}
>{\RaggedRight\arraybackslash}p{0.22\textwidth}
>{\RaggedRight\arraybackslash}p{0.20\textwidth}
>{\RaggedRight\arraybackslash}p{0.21\textwidth}
@{}}
\toprule

\textbf{Survey family} &
\textbf{Scope} &
\textbf{System / agency boundary} &
\textbf{Evidence / deployment} &
\textbf{Gap relative to this review} \\

\midrule

General LMA survey \cite{ref4} &
Cross-domain LMA architectures, memory, planning, tools, action, and
evaluation &
Generic multimodal agents and action mechanisms; no
transportation-specific model/system boundary &
General evaluation frameworks, benchmarks, and research challenges &
No cross-ITS study-level linkage among specialist tools, capability,
validation, and action authority \\

Autonomous-driving MLLM, VLM, and VLA surveys
\cite{ref5,ref6,ref8,ref59} &
Driving perception, reasoning, planning, decision-making, and action &
Direct multimodal processing and perception-to-action architectures;
no common cross-domain C0--C3 distinction &
Driving datasets, benchmarks, open- and closed-loop evaluation,
robustness, safety, and verification &
Vehicle-specific scope; no cross-ITS separation of capability,
validation setting, and specialist authority \\

Broad ITS LLM surveys \cite{ref7,ref60,ref61} &
LLM applications across traffic operations, vehicles, safety, public
transportation, and mobility services &
Multimodal and agent applications are discussed, but model-level,
system-level, and hybrid multimodality are not consistently coded &
Technical performance, ethics, safety, scalability, and deployment
constraints &
No common study-level coding of architecture, feedback, capability,
validation, and action authority \\

System- and role-oriented transportation reviews
\cite{ref9,ref58} &
Transportation-system roles, semantic reasoning, tool coordination, and
research roadmaps &
Heterogeneous evidence and specialist tools are emphasised, but direct
and tool-mediated multimodality are not independently coded &
System frameworks, governance, accountability, and operational
considerations &
No orthogonal family-level classification of architecture, capability,
validation, and proposition evidence \\

Systematic and bibliometric transportation reviews
\cite{ref10,ref11,ref13,ref57} &
Transportation applications, model families, and publication themes &
Multimodal and generative-AI applications are mapped, but pre-agentic,
decision-support, bounded-action, and outcome-responsive roles are not
consistently distinguished &
Performance, efficiency, scalability, safety, governance, and
deployment themes &
No unified source-resolved
architecture--capability--validation--authority map across ITS domains \\

Planning-specific review \cite{ref12} &
Transportation analysis, modelling, system design, and planning &
Multimodality and agency considered within planning rather than as
general ITS system boundaries &
Evaluation, implementation, and decision-support challenges &
Domain-specific scope; no comparison of specialist and LMA authority
across broader ITS applications \\

\textbf{Present review} &
Verified map of 42 foundation-model-centred study families across ITS
operations, driving, safety, public transportation, planning, and
simulation &
Model-level, system-level, and hybrid multimodality; five system
families; C0--C3 capability levels &
Independent E0--E4 validation, P1--P3 proposition evidence, result
direction, Q1--Q8 concerns, comparator strength, robustness, and public
artefacts &
Source/version-resolved linkage among architecture, specialist
baselines, semantic contribution, capability, validation,
methodological concern, and action authority \\

\bottomrule
\end{tabular}

\vspace{2pt}

\begin{minipage}{\textwidth}
\scriptsize
\RaggedRight
\textit{Note:} Survey families are grouped for readability; individual
surveys may differ in methods, terminology, and coverage.
\end{minipage}
}

\end{table*}

\section{Positioning Against Related Surveys}
\label{related-surveys-and-unresolved-review-gap}

The review gap introduced in Section~\ref{introduction} is assessed
against the 15 related surveys in the audited source map. These surveys
differ in scope, unit of analysis, system boundary, and treatment of
empirical evidence and are organised here into three analytical
streams: general LMA surveys, autonomous-driving-focused multimodal and
action-oriented surveys, and broader transportation reviews. This
grouping is for comparison only and does not imply identical methods or
conclusions within each stream.

General LMA surveys establish foundations for multimodal perception,
planning, memory, tool use, action generation, multi-agent
coordination, and evaluation \cite{ref4}. Their cross-domain scope,
however, does not require transportation-specific separation of
semantic interpretation, specialist numerical computation, bounded
action, and outcome-responsive re-decision.

Autonomous-driving surveys of multimodal large language models,
vision--language models, and vision--language--action systems
\cite{ref5,ref6,ref8,ref59} provide detailed coverage of scene
understanding, reasoning, planning, control, datasets, benchmarks,
robustness, and verification. They are the closest prior foundation for
perception-to-action ITS systems, but remain predominantly
vehicle-centred and generally emphasise direct model-level
multimodality rather than a common cross-ITS comparison with
language-centred agents that obtain multimodality through specialist
tools and data systems.

Broader transportation reviews address LLMs, generative AI, and
foundation models across traffic operations, autonomous driving, public
transportation, safety, planning, simulation, and mobility services
\cite{ref7,ref9,ref10,ref11,ref12,ref13,ref57,ref58,ref60,ref61}.
They contribute application taxonomies, technical frameworks,
systematic and bibliometric evidence, resource summaries, governance
considerations, and research roadmaps. The unresolved gap is therefore
not whether multimodality, tool use, feedback, safety, governance, or
deployment have been discussed, but whether they have been jointly
operationalised as reproducible study-level dimensions separating
architecture, capability, validation, proposition evidence,
methodological concern, and action authority.

Table \ref{tab:related-surveys} summaries these differences and the analytical boundary of
the present review.

Prior surveys therefore establish the technical foundations,
application landscape, evaluation challenges, and deployment concerns
of foundation models and agents in transportation. The contribution of
this review is not to introduce previously undiscussed topics, but to
operationalise them jointly within one source-resolved study-level
framework. Within the audited survey set, no prior review was identified
that simultaneously distinguishes model-level from tool-mediated
system-level multimodality; separates pre-agentic, decision-support,
bounded-action, and outcome-responsive capability; evaluates capability
independently from validation setting; separates proposition directness
from result direction and methodological concern; and links these
dimensions to specialist-system boundaries and action authority across
multiple ITS domains. The present review addresses this gap through its
verified architecture--capability--validation--evidence--authority map.

\section{Evidence Across ITS Application Domains}
\label{evidence-across-intelligent-transportation-scenarios}

The 42 primary study families are assigned to five mutually exclusive
application domains according to the principal evaluated role of the
foundation-model-centred system, not paper title. Thus, TransitGPT is
grouped with public operations because it evaluates transportation-data
analysis and decision support, whereas Virtual Traffic Police is grouped
with planning and simulation because its foundation-model contribution
is simulator-mediated controller configuration. Each family is counted
once.

Each domain is examined using the same sequence: transportation task and
specialist baseline; LMA role, architecture, and multimodality;
evaluation data or environment; claim-matched proposition evidence;
capability and validation setting; robustness and failure coverage; and
remaining specialist, operational, or authority boundaries. Because
tasks, datasets, metrics, and experimental designs are heterogeneous,
the synthesis reports verified study-family patterns rather than pooled
effect sizes. The independent meanings of capability, validation,
proposition evidence, methodological concern, and result direction
remain as defined in Sections~\ref{review-methodology} and
\ref{conceptual-framework-of-lmas-in-its}.

\subsection{Prediction and Network Understanding}
\label{traffic-prediction-and-network-state-understanding}

Traffic prediction is dominated by specialist statistical,
spatiotemporal, graph-based, recurrent, and generative models with
explicit objectives, inputs, calibration, and latency requirements.
Foundation models may complement them through cross-source
representation or semantic interpretation. UrbanGPT adapts a
language-model architecture to spatiotemporal urban prediction
\cite{ref14}; TSGDiff provides a contextual multisource generative
baseline \cite{ref15}; and Ye et al. demonstrate heterogeneous
graph-based completion and prediction for metro origin--destination
flows \cite{ref16}.

UrbanGPT is the only primary family whose principal evaluated function
is prediction and network-state understanding. It directly evaluates
P3, provides partial P1 evidence, and does not evaluate P2. Its
demonstrated level is C0/E1: the foundation model contributes to
offline prediction without participating in an actionable
decision-support or action cycle. The claim-matched comparison is
moderate and robustness evidence is partial. The evidence therefore
supports feasibility of a spatiotemporal foundation-model contribution
in one family, but not domain-wide numerical superiority, operational
decision benefit, or deployment readiness.

For stable, high-frequency forecasting, the reviewed evidence does not
displace calibrated specialist models. A stronger ITS-LMA role may
arise when incident narratives, weather, temporary facility conditions,
regulations, or conflicting records alter the interpretation of similar
numerical states. In such settings, the LMA should organize contextual
evidence and formulate the forecasting task, while specialist models
retain numerical estimation, calibration, uncertainty quantification,
and latency-critical execution. Cross-city transfer, sensor failure,
timestamp and map misalignment, event-conditioned calibration, and
downstream decision benefit remain open evaluation requirements.

\subsection{Autonomous Driving and Vehicle Interaction}
\label{autonomous-driving-and-vehicle-road-interaction}

Autonomous-driving applications span model-level multimodal reasoning,
tool-assisted planning, training-time foundation-model support, bounded
program execution, and outcome-responsive agents. Their specialist
baseline is a safety-critical stack of perception, prediction, motion
planning, dynamics, control, and fallback. DriveLM and OmniDrive
evaluate relation-aware or counterfactual scene reasoning and open-loop
planning \cite{ref17,ref20}; Dolphins and DriveGPT4 extend model-level
driving understanding \cite{ref63,ref82}; and LLaDA adapts semantic
plans to jurisdiction-specific rules without runtime vehicle authority
\cite{ref64}. LaMPilot generates programs that invoke bounded driving
primitives in simulation and is therefore C2/E2 rather than C3
\cite{ref65}.

Agent-Driver, LMDrive, DiLu, DriveVLM, DriveGPT4-V2, ORION, and
MindDriver use updated simulated observations, memory, or
specialist-pipeline outputs to condition later driving decisions
\cite{ref18,ref27,ref62,ref69,ref70,ref71,ref74}. These families
demonstrate C3 functionality, but remain primarily at E2 interactive
simulation. By contrast, VLM-RL is C0/E2 because the vision--language
model supplies a training-time semantic reward while a separate
reinforcement-learning policy executes CARLA actions \cite{ref19};
V2X-UniPool is C0/E1 because its cooperative knowledge-pool and
reasoning evaluation remains offline \cite{ref75}. These cases
illustrate why capability is assigned to the demonstrated
foundation-model role rather than inherited from downstream execution.

Only one family reaches E3. The zero-shot LLM-guided driving study
evaluates a multimodal foundation-model-centred system in controlled
real-world closed-loop experiments, with periodically processed sensor
evidence producing high-level instructions for a low-level controller
\cite{ref84}. This exceeds simulation evidence but does not establish
E4 sustained deployment because longitudinal safety outcomes, broad
operational-design-domain coverage, incident rates, independent
operational audit, and sustained routine authority are not reported.

Across the 17 autonomous-driving families, P1 is direct in 12 and
partial in five, while P3 is direct in 15 and partial in two. Sixteen
families have strong or moderate claim-matched comparisons, providing
the strongest domain-level evidence for semantic scene interpretation
and multidimensional integration. Evidence for reconciliation and
deployment is substantially weaker: P2 is partial in four and not
evaluated in 13; robustness is partial in nine and inadequate in eight;
and Q8 is high concern in 11. Seven families use CARLA as the primary
environment, six are primarily offline, two are primarily real-world or
onboard, and the remainder use other simulated settings.

The best-supported LMA roles are semantic scene interpretation, intent
and rule grounding, high-level planning, explanation, memory-supported
reasoning, and specialist-tool coordination. Dedicated prediction and
motion-planning systems, such as the risk-aware contingency-planning
approach in RACP \cite{ref21}, should retain explicit dynamics and
safety constraints. Direct task evidence for an LMA component therefore
does not justify unconstrained low-level control, unrestricted
operational authority, or replacement of independently verifiable
planning, control, and fallback functions.

\subsection{Signal Control and Public Operations}
\label{traffic-signal-control-and-public-infrastructure-operations}

Signal control combines strict phase, clearance, coordination, demand,
and safety constraints with changing traffic conditions and operational
priorities. Operational authority therefore remains with established
fixed-time or actuated logic, optimisation, max-pressure control, or
multi-agent reinforcement learning. The reviewed LMA families instead
add language interpretation, high-level decision-making, or
tool-mediated access to operational data.

LLMLight conditions phase decisions on updated CityFlow states
\cite{ref28}, while Movahedi and Choi use knowledge from SUMO
traffic-control interactions to inform later phase selection
\cite{ref29}. Both are C3/E2 because transportation-state feedback
changes subsequent foundation-model decisions in simulation. Masri
et al. evaluate conflict, priority, and driver-guidance outputs
offline and remain C1/E1 \cite{ref30}. TransitGPT analyses GTFS data
through generated Python while implementation authority remains with
the user, and is therefore C1/E1 \cite{ref39}. Its retry-enabled
configuration increased task accuracy from 74\% to 90\% for GPT-4o and
from 84\% to 93\% for Claude-3.5-Sonnet; end-to-end latency decreased
from 20.7~s to 17.2~s for GPT-4o but increased from 12.5~s to 18.5~s
for Claude-3.5-Sonnet. This demonstrates a tool-use
accuracy--latency trade-off, not an operational transportation outcome.

Across the four families, P1 and P3 are partial in every case. P2 is
partial only for TransitGPT and not evaluated in the other three. All
four claim-matched comparisons are partial or nonisolating; robustness
is partial in two families and inadequate in two; and none reaches E3
or E4. Thus, the evidence supports language-mediated data access,
incident and policy interpretation, strategy suggestion, explanation,
and bounded controller configuration, but not LMA-specific control
superiority.

Optimisation, max-pressure control, and MARL remain the appropriate
specialist baselines \cite{ref31,ref32}. Phase feasibility, timing,
coordination, hard constraints, safe execution, independent monitoring,
fallback, and rollback should remain with verified controllers and
accountable operators until matched comparisons and controlled
operational evidence justify greater LMA authority.

\subsection{Traffic Safety and Incident Management}
\label{traffic-safety-incident-analysis-and-risk-communication}

Traffic-safety applications include event detection and localisation,
accident-scene understanding, responsibility-support reasoning,
safety-critical scenario generation, and incident-management decision
support. Specialist baselines include validated detection models,
calibrated risk and survival models, causal analysis, optimisation,
emergency procedures, and accountable human adjudication. LMAs may
integrate visual, temporal, textual, legal, and operational evidence,
but appropriate authority depends on the task.

Abu Tami et al., VRU-Accident, SafePLUG, and AITP evaluate multimodal
event detection, dense captioning, pixel- or temporal-level grounding,
or responsibility-allocation support
\cite{ref34,ref35,ref36,ref73}. These are primarily C0/E1 because they
provide perception, grounding, or analytical outputs without an action
cycle. ChatScene translates language descriptions into executable
CARLA safety scenarios and is C2/E2 \cite{ref66}, whereas LLMScenario
remains C0/E1 because its score-guided generation process does not
execute a transportation action \cite{ref68}. The highway
incident-management family combines language reasoning with specialist
optimisation over historical incident data, while implementation
remains with human operators, yielding C1/E1 \cite{ref85}.

Among the seven families, four directly evaluate P1 and four directly
evaluate P3; the remaining three are partial for each. P2 is partial in
five and not evaluated in two. Four families have strong or moderate
claim-matched comparisons, but six remain at E1 and none reaches E3 or
E4. The evidence therefore supports event interpretation, evidence
localisation, structured description, scenario generation, explanation,
and analyst triage, but not prospective safety improvement from an
LMA-mediated intervention.

Structured causal, survival, and calibrated risk models remain
necessary for estimating risk and intervention effects. Text-only
crash-narrative analysis is retained as contextual evidence of both
accessibility and reasoning limitations \cite{ref37}. LMA outputs
should therefore remain traceable analytical hypotheses rather than
automatic causal conclusions, legal determinations, enforcement
decisions, or emergency authority. Prospective external validation,
calibration, false-alert burden, subgroup performance, privacy,
source attribution, rare-event coverage, and measured operator benefit
remain necessary before greater decision authority is justified.

\begin{table*}[!t]
\centering
\caption{Verified application-domain evidence profile for the 42 primary
study families.}
\label{tab:application-domain-evidence}

{\footnotesize
\setlength{\tabcolsep}{2.5pt}
\renewcommand{\arraystretch}{1.10}

\begin{tabular}{@{}
>{\RaggedRight\arraybackslash}p{0.16\textwidth}
>{\centering\arraybackslash}p{0.04\textwidth}
>{\centering\arraybackslash}p{0.115\textwidth}
>{\centering\arraybackslash}p{0.115\textwidth}
>{\RaggedRight\arraybackslash}p{0.18\textwidth}
>{\RaggedRight\arraybackslash}p{0.31\textwidth}
@{}}

\toprule

\textbf{Application domain} &
\textbf{$n$} &
\textbf{C0--C3} &
\textbf{E1--E4} &
\textbf{P1--P3 (D/P/N)} &
\textbf{Comparator / robustness / boundary} \\

\midrule

Prediction and network understanding &
1 &
1 / 0 / 0 / 0 &
1 / 0 / 0 / 0 &
\shortstack[l]{P1: 0/1/0\\P2: 0/0/1\\P3: 1/0/0} &
One moderate comparison and partial robustness; single-family evidence
does not establish forecasting superiority or operational readiness. \\

Autonomous driving and vehicle interaction &
17 &
6 / 2 / 1 / 8 &
7 / 9 / 1 / 0 &
\shortstack[l]{P1: 12/5/0\\P2: 0/4/13\\P3: 15/2/0} &
Ten strong, six moderate, and one partial comparison; robustness is
partial in nine and inadequate in eight; one E3 family and no E4
safety case. \\

Signal control and public operations &
4 &
0 / 2 / 0 / 2 &
2 / 2 / 0 / 0 &
\shortstack[l]{P1: 0/4/0\\P2: 0/1/3\\P3: 0/4/0} &
All comparisons are partial or nonisolating; no direct P1 or P3
contribution and no controlled operational evidence. \\

Traffic safety and incident management &
7 &
5 / 1 / 1 / 0 &
6 / 1 / 0 / 0 &
\shortstack[l]{P1: 4/3/0\\P2: 0/5/2\\P3: 4/3/0} &
Four strong or moderate comparisons, but predominantly offline
validation; analytical assistance does not justify legal or
operational authority. \\

Planning, simulation, and HMC &
13 &
0 / 3 / 6 / 4 &
5 / 8 / 0 / 0 &
\shortstack[l]{P1: 7/6/0\\P2: 0/10/3\\P3: 4/9/0} &
Six strong, one moderate, and six partial comparisons; P3 is partial in
nine and no family reaches E3 or E4. \\

\bottomrule

\end{tabular}
}

\end{table*}

\subsection{Planning, Simulation, and Human--Machine Collaboration}
\label{emerging-directions-planning-simulation-and-human-machine-collaboration}

The 13-family planning, simulation, modelling, and human--machine
collaboration (HMC) domain contains the widest range of evaluated
system forms. Specialist authority in this domain resides in
optimisation, assignment, discrete-choice and behavioural models,
network loading, calibrated simulation, feasibility checking, and
human approval. LMA contributions include preference interpretation,
scenario authoring, tool orchestration, model enhancement, plan
explanation, and feedback-supported replanning.

Four families demonstrate C3/E2. LLMTraveler updates route-choice
memory from experienced travel times \cite{ref38}; GATSim represents
mobility agents whose experience and environmental change affect later
plans \cite{ref72}; the MATSim replanning agent uses simulator
feedback and a verifier to revise electric-vehicle charging plans
\cite{ref77}; and ChatSUMO-Agent uses state-aware tools and SUMO
feedback for subsequent replanning \cite{ref83}. These systems
demonstrate outcome-responsive behaviour in virtual transportation
environments, not operational authority over a real network.

Six families demonstrate bounded C2 functionality: Speak to Simulate
and the AI Research Agent \cite{ref41,ref42}; collaborative
LLM-based editable-scene simulation \cite{ref67}; human-approved
traffic-simulation orchestration \cite{ref78}; image--text-guided
procedural scene generation \cite{ref81}; and Virtual Traffic Police
\cite{ref80}. Virtual Traffic Police is assigned to this domain
because the evaluated foundation-model contribution configures
specialist controllers in simulation for unforeseen incidents; the
specialist adaptive controllers retain the control function.
ChatSUMO, parking-planning support, and multimodal e-mobility
optimisation remain C1/E1 because an external user validates or
implements their outputs \cite{ref40,ref76,ref79}. ProSim is retained
only as a contextual specialist simulator because the foundation model
is not the centre of the complete simulation architecture
\cite{ref90}.

P1 is directly evaluated in seven of the 13 families and partially
evaluated in six. P2 is partial in ten and not evaluated in three. P3
is direct in four and partial in nine, showing that tool-rich or
multimodal workflows frequently demonstrate integration without
isolating its contribution. Six families have strong claim-matched
comparisons, one has a moderate comparison, and six have partial or
nonisolating comparisons. Robustness is partial in ten and inadequate
in three; none reaches E3 or E4.

The evidence supports preference elicitation, scenario and artefact
generation, simulation setup, data and tool coordination,
feedback-supported virtual replanning, explanation, and
human-in-the-loop assistance. It does not establish that an
LMA-generated plan is feasible, optimal, behaviourally realistic, or
operationally deployable. Optimisation, demand and route-choice
models, network loading, simulation calibration, physical and policy
constraints, validation of generated artefacts, and final execution
should remain with specialist systems and accountable users.

Table \ref{tab:application-domain-evidence} summarises the verified quantitative evidence profile across
the five application domains. Capability, validation setting,
proposition evidence, comparator strength, robustness, and claim
boundaries are reported separately rather than combined into a
composite score.

Across the corpus, demonstrated capability exceeds validation maturity,
and increasing architectural complexity or delegated execution does not
by itself establish operational readiness. Direct P1 or P3 evidence is
supported only where claim-matched comparisons are strong or moderate,
whereas P2 remains unresolved, with no family directly evaluating the
complete provenance--challenge--handling--comparison--outcome chain.
Table \ref{Cross-domain_evidence} translates these cross-domain patterns into the best-supported
LMA roles, specialist-system boundaries, and bounded inferences for each
application domain.

\begin{table*}[!t]
\centering
\caption{Cross-domain LMA roles, specialist boundaries, and bounded
inferences.}
\label{Cross-domain_evidence}

{\footnotesize
\setlength{\tabcolsep}{2.5pt}
\renewcommand{\arraystretch}{1.10}

\begin{tabular}{@{}
>{\RaggedRight\arraybackslash}p{(\textwidth - 10\tabcolsep) * \real{0.14}}
>{\RaggedRight\arraybackslash}p{(\textwidth - 10\tabcolsep) * \real{0.20}}
>{\RaggedRight\arraybackslash}p{(\textwidth - 10\tabcolsep) * \real{0.22}}
>{\RaggedRight\arraybackslash}p{(\textwidth - 10\tabcolsep) * \real{0.16}}
>{\RaggedRight\arraybackslash}p{(\textwidth - 10\tabcolsep) * \real{0.13}}
>{\RaggedRight\arraybackslash}p{(\textwidth - 10\tabcolsep) * \real{0.15}}
@{}}

\toprule
\textbf{Domain} &
\textbf{Specialist authority} &
\textbf{Supported LMA role} &
\textbf{Proposition evidence} &
\textbf{Highest C/E} &
\textbf{Bounded inference} \\
\midrule

Prediction and network understanding &
Calibrated graph and spatiotemporal forecasting &
Contextual interpretation, cross-source querying, and explanation
around a matched numerical forecaster &
P1-P; P2-N; P3-D &
C0/E1 &
One family supports integration feasibility, not domain-wide numerical
superiority. \\

Autonomous driving and vehicle interaction &
Verified perception, prediction, motion planning, control, and safety
fallback &
Semantic scene reasoning, intent and rule grounding, high-level
planning, explanation, memory, and tool coordination &
P1: 12 D; P3: 15 D; P2: 4 P &
C3/E3 &
Direct task evidence does not establish sustained autonomous authority
or an E4 safety case. \\

Signal control and public operations &
Optimisation, max-pressure or MARL control, phase feasibility, and safe
execution &
Natural-language access, incident interpretation, strategy suggestion,
explanation, and controller configuration &
P1-P and P3-P in all four families &
C3/E2 &
Workflow feasibility is not LMA-specific control superiority or field
readiness. \\

Traffic safety and incident management &
Validated detection, causal and risk models, emergency procedures, and
accountable adjudication &
Event interpretation, evidence localisation, scenario generation,
explanation, and analyst support &
P1: 4 D; P3: 4 D; P2: 5 P &
C2/E2 &
Analytical assistance does not justify legal, enforcement, emergency,
or autonomous incident-response authority. \\

Planning, simulation, and HMC &
Optimisation, assignment, behavioural models, network loading,
calibrated simulators, and human approval &
Preference translation, scenario authoring, simulation setup, tool
orchestration, virtual replanning, and plan explanation &
P1: 7 D; P3: 4 D; P2: 10 P &
C3/E2 &
Workflow automation does not establish feasibility, optimality,
behavioural validity, simulation validity, or deployment maturity. \\

\bottomrule
\end{tabular}
}

\end{table*}

\section{Cross-Study Synthesis, Assurance, and Deployment Roadmap}
\label{cross-scenario-limitations-resources-and-deployment-roadmap}

Section~\ref{evidence-across-intelligent-transportation-scenarios}
established the domain-level evidence profile. This section synthesises
the cross-study limitations, reproducibility conditions, assurance
requirements, and evaluation steps that determine how far the reported
capabilities can support transportation decisions. The synthesis
preserves the independence of capability, validation setting,
proposition evidence, methodological concern, and result direction;
absence of a particular evaluation is treated as an evidence gap rather
than proof of failure or impossibility.

\subsection{Limitations and Assurance}
\label{cross-scenario-limitations-and-safeguards}

The evidence base remains concentrated in autonomous driving and
planning, simulation, and human--machine collaboration, while
evaluation is dominated by offline and simulated settings. Only two
families were evaluated primarily in real-world or onboard settings,
so current evidence does not establish broad operational
representativeness. Detailed evaluation-environment profiles are
provided in Supplementary Table~S7.

Five technical limitations recur across domains. First, multimodal
grounding can fail under occlusion, adverse weather, distribution shift,
stale maps, temporal or spatial misalignment, and adversarial text or
images. Second, fluent semantic completion can conceal disagreement,
drop source qualifications, or construct a plausible but false
transportation state. Third, autoregressive reasoning does not replace
causal identification, numerical optimisation, probabilistic
calibration, physical feasibility, or formal safety constraints.
Fourth, cloud-scale inference and tool orchestration may violate
latency, reliability, energy, communication, and data-residency
requirements. Fifth, agentic tool use introduces prompt injection,
unauthorised action, stale or contaminated memory, privacy leakage, and
unclear responsibility. Indirect prompt injection is especially
relevant when an agent retrieves untrusted reports or external content
before calling tools \cite{ref43}.

These limitations make assurance an architectural rather than purely
model-level requirement. For every evaluated or deployed workflow, the
system boundary should identify the executor, action authority,
operational envelope, data provenance, uncertainty representation,
human role, independent monitor, fallback controller, rollback
mechanism, audit record, and post-deployment monitoring responsibility.
A generated recommendation should not be treated as evidence of its own
correctness, and an LMA should not verify a safety-critical action using
only the same reasoning process that produced it.

Comparator and robustness evidence constrain the strength of current
claims. Nineteen families have strong claim-matched comparison
evidence, nine have moderate evidence, and 14 have only partial or
nonisolating comparisons. All direct P1 and P3 judgments occur in the
strong or moderate groups. However, no family provides substantive
robustness and failure evidence under Q6: 28 provide partial
boundary-condition evidence and 14 remain inadequate. Q5
setting--claim alignment is low concern in one family, some concern in
37, and high concern in four. Q8 authority, safety, governance, and
deployment readiness is some concern in 18 families, high concern in
18, and not applicable in six genuinely non-action offline families;
11 of the 14 C3 families have high Q8 concern. These are
claim-specific evidence profiles, not study rankings.

Established standards provide engineering anchors but do not validate
an ITS-LMA by themselves. ISO 26262 supports the road-vehicle
functional-safety lifecycle \cite{ref44}; ISO 21448 addresses hazards
arising from intended functionality and performance limitations
\cite{ref45}; ISO/SAE 21434 addresses road-vehicle cybersecurity
engineering \cite{ref46}; ISO/PAS 8800 addresses safety and artificial
intelligence in road vehicles \cite{ref47}; and IEEE 7001 addresses
transparency of autonomous systems \cite{ref48}. For network
operations, planning, and non-vehicle applications, the NIST AI Risk
Management Framework \cite{ref49}, ISO/IEC 23894 \cite{ref50}, and
ISO/IEC 42001 \cite{ref51} provide governance and organisational
risk-management anchors. These sources must be translated into
scenario-specific requirements and cannot substitute for
claim-matched validation, demonstrated operational benefit, or a
system-level safety and accountability case.

\subsection{Resources and Reproducibility}
\label{resources-and-evaluation-agenda}

Frequently reused resources include nuScenes \cite{ref52}, the Waymo
Open Dataset \cite{ref53}, CityFlow \cite{ref55}, and SUMO
\cite{ref56}; highD \cite{ref54} and DriveCombo \cite{ref91} provide
contextual resources for trajectory behaviour and compositional rule
reasoning. Reuse of a named resource does not make studies directly
comparable because geography, sensing, scenario coverage, splits,
prompts, model versions, and operational realism may differ.

Public implementation code was located for 21 families, study-specific
data for 11, and prompts or configurations for six; only four families
expose all three major artefact types. Detailed evaluation-environment
and public-artefact profiles are provided in Supplementary Table~S7.

The principal evaluation need is therefore not another static benchmark
alone, but a benchmark-to-case programme that tests source conflict,
sensor outage, rare events, distribution shift, cross-city transfer,
operator interaction, latency, tool failure, and verified
transportation outcomes. Agentic refinement of established models is a
promising near-term route because it can preserve specialist structure
while making the generated changes inspectable. The AI Research Agent,
for example, generates, executes, evaluates, and iteratively refines
transportation-model code on fixed datasets \cite{ref42}; other systems
use verifier-gated replanning or human-approved simulation toolchains
\cite{ref77,ref78}. Reproducible evaluation should retain executable
artefacts, input provenance, dataset and simulator versions, model and
tool versions, generated code, prompts or configurations, validation
outputs, failure logs, and human approvals.

\subsection{Answers to the Research Questions}
\label{answers-to-the-research-questions}

\textbf{RQ1.} Measurable LMA value is best supported when claim-matched
comparisons isolate the contribution of transportation semantics or the
integration of distinct evidence dimensions. The current evidence
supports task-specific benefits for P1 and P3, whereas P2 remains
unresolved because no family directly evaluates the complete
provenance--challenge--handling--comparison--outcome chain. The evidence
therefore supports targeted semantic and integration benefits rather
than general superiority over specialist transportation models.

\textbf{RQ2.} The reviewed ITS-LMA systems are technically heterogeneous
and increasingly tool-mediated, spanning direct foundation-model
inference, specialist-model pipelines, and agentic orchestration with
model-level, system-level, or hybrid multimodality. Demonstrated
capability substantially exceeds validation maturity: C2 and C3
functionality is increasingly evident, but validation remains
concentrated at E1 and E2, with only one E3 family and no E4 evidence.
Agency, multimodality, and delegated execution therefore indicate
architectural capability, not robustness, operational maturity, or
justified autonomous authority.

\textbf{RQ3.} The current evidence supports an LMA-orchestrator and
specialist-executor architecture. LMAs are best supported for semantic
interpretation, intent and preference translation, evidence
organisation, explanation, scenario authoring, task decomposition, and
specialist-tool coordination, while forecasting, optimisation,
simulation, hard constraints, low-level control, safety fallback, legal
judgment, and final operational authority should remain with validated
specialist systems or accountable humans. Greater LMA authority would
require stronger proposition-specific evidence, reliable
outcome-responsive operation, traceable evidence reconciliation, robust
failure handling, and controlled operational benefit.

\textbf{RQ4.} The principal barriers to greater LMA authority are
unresolved evidence reconciliation, inadequate robustness and external
validation, fragmented reproducibility, and governance that has not yet
matched agentic capability. No family demonstrates E4 sustained
deployment. Progress therefore requires matched comparative testing,
failure and cyber-resilience evaluation, controlled E3 trials,
independent monitoring and fallback, transparent reproducibility, and
longitudinal E4 evidence within a declared operational envelope.

\begin{table*}[!t]
\centering
\caption{Domain-specific evaluation priorities and next authority gates.}
\label{Domain-specific evaluation}

{\footnotesize
\setlength{\tabcolsep}{2.5pt}
\renewcommand{\arraystretch}{1.10}

\begin{tabular}{@{}
>{\raggedright\arraybackslash}p{(\textwidth - 8\tabcolsep) * \real{0.14}}
>{\raggedright\arraybackslash}p{(\textwidth - 8\tabcolsep) * \real{0.22}}
>{\raggedright\arraybackslash}p{(\textwidth - 8\tabcolsep) * \real{0.24}}
>{\raggedright\arraybackslash}p{(\textwidth - 8\tabcolsep) * \real{0.20}}
>{\raggedright\arraybackslash}p{(\textwidth - 8\tabcolsep) * \real{0.20}}@{}}
\toprule
\textbf{Domain} &
\textbf{Evidence bottleneck} &
\textbf{Priority test} &
\textbf{Assurance evidence} &
\textbf{Next authority gate} \\
\midrule

Prediction and network understanding &
One C0/E1 family; P3-D but no operational decision-benefit evidence &
Same specialist forecaster alone, LMA alone, and combined across
cities and event conditions &
Calibration, drift, source conflict, sensor outage, latency, and
downstream decision-effect analysis &
Controlled decision support only after matched benefit and reliability \\

Autonomous driving and vehicle interaction &
Strong P1/P3 evidence but no direct P2 case, limited robustness, one E3,
and no E4 &
Matched specialist stack, LMA-only condition, LMA-orchestrated stack,
and complete monitored architecture within a declared ODD &
Intervention and fallback records, adversarial and rare-event tests,
cyber-resilience, calibration, and longitudinal safety outcomes &
Restricted E3 authority before any E4 claim; retain independent planning,
control, and safety fallback \\

Signal control and public operations &
P1 and P3 remain partial in all four families; no controlled field
evidence &
Matched specialist controller with and without LMA interpretation,
configuration, and assurance components across networks &
Constraint violations, transfer, latency, rollback, fairness, operator
workload, and failure recovery &
Retain execution with verified controllers until supervised field
evidence demonstrates incremental operational benefit \\

Traffic safety and incident management &
Predominantly offline evidence; no prospective intervention benefit or
operational validation &
Specialist or analyst workflow alone versus the same workflow with LMA
grounding and explanation under blinded prospective evaluation &
Calibration, false-alert burden, subgroup performance, provenance,
privacy, intervention effect, and accountable adjudication &
Advisory use only until prospective benefit and responsibility
boundaries are demonstrated \\

Planning, simulation, and HMC &
No E3; P3 remains partial in nine families; feasibility and behavioural
validity are often untested &
Specialist workflow alone versus LMA-only generation versus
LMA-orchestrated and verifier-gated workflows &
Executable correctness, local calibration, behavioural realism,
distributional impact, tool failure, and human approval records &
Human-approved E2 workflows before bounded operational decision support;
no autonomous execution from workflow automation alone \\

\bottomrule
\end{tabular}
}

\end{table*}

\subsection{Comparative Evaluation and Roadmap}
\label{recommended-comparative-evaluation-protocol}

Future experiments should compare four matched configurations under the
same data, scenarios, specialist model, action space, operational
constraints, and evaluation budget: (A) the specialist transportation
system alone; (B) an LMA without specialist tools; (C) an LMA
orchestrating the same specialist system; and (D) the complete
architecture with permission gateway, independent monitor, fallback,
rollback, and, where appropriate, human approval. This design separates
the contribution of semantic reasoning from the contribution of the
specialist model and from the additional benefit of the assurance
architecture.

Ablations should remove provenance, uncertainty representation, memory,
tool-output verification, and environmental feedback separately.
Proposition claims require additional claim-matched contrasts:
semantic-element ablation or controlled semantic perturbation for P1;
provenance-aware missingness, uncertainty, or conflict handling versus
an untreated condition for P2; and reduced-input, unimodal,
nonintegrated, misaligned, factorial, or component conditions for P3.
Whole-system superiority, additional data, model size, or tool access
should be reported as potential confounding rather than attributed
automatically to a proposition. A C3 claim requires observed
transportation outcomes to enter the foundation-model state, context,
or memory and demonstrably alter a subsequent model-level decision;
software retry or a sequence of simulator observations alone is
insufficient.

Reporting should cover task performance and calibration together with
end-to-end latency, token and compute cost, energy where measurable,
tool-call validity, code or scenario execution failure, rule violation,
unsafe-action rate, human intervention, fallback activation, rollback
success, distribution shift, sensor outage, adversarial input, and
cross-network transfer. Statistical uncertainty, repeated runs,
negative and null findings, model and tool versions, prompts,
configuration files, executable artefacts, and human approvals should
be disclosed. Unreported latency, cost, energy, or failure rates should
remain explicitly ``not reported'' rather than inferred.

\label{staged-roadmap}
The deployment roadmap should progress by demonstrated evidence rather
than by model scale or architectural complexity. Near-term operational
use should remain read-only or advisory, with source attribution,
explicit uncertainty, source-conflict display, verified tool outputs,
and accountable human review. Medium-term work should evaluate bounded
C2/E2 virtual actions under pre-authorisation, independent monitoring,
rollback, cyber-resilience, and structured rare-event and tool-failure
tests. C3 should be claimed only when transportation outcomes reach the
foundation-model decision layer and change a later decision within a
declared operational envelope. Progression to E3 requires controlled
real-world evaluation with intervention, fallback, latency, and failure
records. E4 requires sustained longitudinal operation, incident
reporting, governance, independent audit, and evidence that the
complete system remains within its authorised envelope. Across all
stages, P1 and P3 require claim-matched contribution tests, while P2
requires traceable provenance and controlled evidence-quality
challenges.

Table~\ref{Domain-specific evaluation} translates these general
principles into domain-specific evaluation and authority gates.
Unlike Table~\ref{Cross-domain_evidence}, which summaries the current
division of responsibility, it focuses on the next evidence needed to
justify progression.

\section{Limitations}
\label{limitations}

Four limitations define the scope of interpretation. First, construct
validity is constrained by inconsistent use of terms such as
\emph{large multimodal agent}, \emph{agent}, \emph{multimodality},
\emph{closed loop}, \emph{deployment}, and \emph{real-world
validation}. The operational definition, foundation-model-centred
system boundary, independent capability and validation scales, and
conservative lower-code rule reduce, but cannot eliminate, this
ambiguity.

Second, selection and coding validity are limited because eligibility,
source resolution, extraction, and analytical coding were conducted
within one author-led process with structured model assistance.
Prespecified manuals and label-masked within-process audits reduce
unrecorded inconsistency, but they do not constitute independent
duplicate human coding; no inter-rater reliability statistic is claimed.
Detailed stability results are provided in Supplementary Table~S5.

Third, external validity is limited by reliance on publicly accessible
English-language reports, rapid publication and version turnover, and
likely under-representation of private industrial evaluations. The
corpus is concentrated in autonomous driving (17 families) and
planning, simulation, and human--machine collaboration (13), whereas
prediction and network understanding contains one family. Geography,
demographics, operating conditions, and dataset provenance are
insufficiently reported for country-, population-, or network-level
representativeness. The five application domains are analytical
groupings based on principal evaluated function; alternative defensible
assignments could change domain subtotals without changing the
42-family corpus or family-level C/E/P/Q judgments.

Fourth, conclusion validity is constrained by heterogeneous tasks,
datasets, simulators, metrics, baselines, prompts, model versions, tool
configurations, and reporting practices, precluding statistical
meta-analysis and direct cross-domain effect-size comparison. Latency,
compute and energy cost, calibration, tool failure, unsafe actions,
negative findings, and human interventions are also reported
inconsistently. ``Not located'' denotes that no official public
artefact was identified in the searched principal paper, project page,
or repository, not that no private or later artefact exists.
Database-specific retrieval and deduplication totals were not preserved
during the review's early iterative development; the protocol is
therefore not presented as PRISMA compliant.

Prespecified conservative sensitivity analyses did not alter the central
bounded-orchestration conclusion; even under the most conservative
capability and validation assumptions, bounded-action evidence remains
while sustained-deployment evidence remains absent. Full results are
provided in Supplementary Table~S6. Future updates should also use a
living evidence registry preserving search records, study-family links,
source versions, eligibility decisions, evidence locators, datasets,
prompts, model and tool versions, executable artefacts, failures,
authority boundaries, and post-deployment incidents.

\section{Conclusion}
\label{conclusion}

The synthesis of 42 study families supports a bounded role for LMAs in
ITS: direct evidence is strongest for transportation semantics and
multidimensional integration, while no family directly evaluates the
P2 evidence-reconciliation proposition under the complete
provenance--challenge--handling--comparison--outcome criterion.
Capability also outpaces validation: 14 families reach C3, but only one
reaches E3 and none reaches E4. These findings support
contribution-specific use rather than general replacement or
deployment-ready autonomy.

Current evidence most strongly supports LMAs for semantic
interpretation, intent and preference translation, evidence
organisation, scenario authoring, explanation, task decomposition, and
specialist-tool coordination. Numerical forecasting and calibration,
optimisation and feasibility, behavioural and network simulation,
physical constraints, low-level control, safety fallback, legal
judgment, and final operational authority should remain with
independently validated specialist systems or accountable humans.

Greater LMA authority requires direct evidence-reconciliation and
claim-matched contribution tests, provenance-aware uncertainty handling,
robust failure and cyber-resilience evaluation, controlled E3 trials
with independent monitoring and fallback, reproducible artefacts and
configurations, and longitudinal E4 evidence within declared
operational envelopes. Bounded orchestration is therefore the most
defensible near-term architecture: LMAs can improve how heterogeneous
transportation evidence and specialist tools are interpreted and
coordinated, provided that authority remains explicit and every
safety- or correctness-critical output is independently verifiable.


\begin{thebibliography}{99}

\bibitem{ref1} M. J. Page et al., ``The PRISMA 2020 Statement: An Updated Guideline for Reporting Systematic Reviews,'' BMJ, vol. 372, art. n71, 2021, doi: 10.1136/bmj.n71.

\bibitem{ref2} SAE International, ``Taxonomy and Definitions for Terms Related to Driving Automation Systems for On-Road Motor Vehicles,'' SAE J3016\_202104, 2021.

\bibitem{ref3} Z. Liu et al., ``TRIP: Transport Reasoning With Intelligence Progression---A Foundation Framework,'' Transportation Research Part C: Emerging Technologies, vol. 179, art. 105260, 2025, doi: 10.1016/j.trc.2025.105260.

\bibitem{ref4}
J. Xie, Z. Chen, R. Zhang, and G. Li,
``Large multimodal agents: A survey,''
\emph{Visual Intelligence}, vol. 3, art. 24, 2025,
doi: 10.1007/s44267-025-00093-y.

\bibitem{ref5} C. Cui et al., ``A Survey on Multimodal Large Language Models for Autonomous Driving,'' in Proc. IEEE/CVF WACV Workshops, 2024, pp. 958--979, doi: 10.1109/WACVW60836.2024.00106.

\bibitem{ref6}
S. Jiang et al.,
``A survey on vision--language--action models for autonomous driving,''
in \emph{Proc. IEEE/CVF ICCV Workshops}, 2025, pp. 4583--4595,
doi: 10.1109/ICCVW69036.2025.00476.


\bibitem{ref7} S. Wandelt, C. Zheng, S. Wang, Y. Liu, and X. Sun, ``Large Language Models for Intelligent Transportation: A Review of the State of the Art and Challenges,'' Applied Sciences, vol. 14, no. 17, art. 7455, 2024, doi: 10.3390/app14177455.

\bibitem{ref8}
X. Zhou, M. Liu, E. Yurtsever, B. L. Zagar, W. Zimmer, H. Cao,
and A. C. Knoll,
``Vision language models in autonomous driving: A survey and outlook,''
\emph{IEEE Transactions on Intelligent Vehicles}, early access,
pp. 1--20, 2024,
doi: 10.1109/TIV.2024.3402136.

\bibitem{ref9} T. Nie, J. Sun, and W. Ma, ``Exploring the Roles of Large Language Models in Reshaping Transportation Systems: A Survey, Framework, and Roadmap,'' Artificial Intelligence for Transportation, vol. 1, art. 100003, 2025, doi: 10.1016/j.ait.2025.100003.

\bibitem{ref10} N. Maksoud, H. AlJassmi, L. Ali, and A. R. Masoud, ``Applications of Large Language Models and Generative AI in Transportation: A Systematic Review and Bibliometric Analysis,'' Transportation Research Interdisciplinary Perspectives, vol. 34, art. 101699, 2025, doi: 10.1016/j.trip.2025.101699.

\bibitem{ref11}
Y. Yan et al.,
``Large language models for transportation research: Methodologies,
state of the art, and future opportunities,''
\emph{Information Fusion}, vol. 136, art. 104546, 2026,
doi: 10.1016/j.inffus.2026.104546.

\bibitem{ref12}
Y. Jin and J. Ma,
``A survey of large language models in transportation planning:
Modelling, design and decision-making,''
\emph{Transportmetrica A: Transport Science}, early access,
pp. 1--46, 2026,
doi: 10.1080/23249935.2026.2631152.

\bibitem{ref13}
S. Kaur et al.,
``Harnessing large language models for intelligent transportation
systems: A systematic review,''
\emph{Multimodal Transportation}, vol. 5, no. 3, art. 100308, 2026,
doi: 10.1016/j.multra.2026.100308.

\bibitem{ref14} Z. Li et al., ``UrbanGPT: Spatio-Temporal Large Language Models,'' in Proc. 30th ACM SIGKDD Conf. Knowledge Discovery and Data Mining, 2024, pp. 5351--5362, doi: 10.1145/3637528.3671578.

\bibitem{ref15} H. Zhang, H. Dong, and Z. Yang, ``TSGDiff: Traffic State Generative Diffusion Model Using Multi-Source Information Fusion,'' Transportation Research Part C: Emerging Technologies, vol. 174, art. 105081, 2025, doi: 10.1016/j.trc.2025.105081.

\bibitem{ref16} J. Ye, J. Zhao, F. Zheng, and C.-Z. Xu, ``A Heterogeneous Graph Convolution Based Method for Short-Term OD Flow Completion and Prediction in a Metro System,'' IEEE Transactions on Intelligent Transportation Systems, vol. 25, no. 5, pp. 4488--4500, 2024, doi: 10.1109/TITS.2023.3323756.

\bibitem{ref17} C. Sima et al., ``DriveLM: Driving with Graph Visual Question Answering,'' in Computer Vision---ECCV 2024, 2024, pp. 256--274, doi: 10.1007/978-3-031-72943-0\_15.

\bibitem{ref18} J. Mao, J. Ye, Y. Qian, M. Pavone, and Y. Wang, ``A Language Agent for Autonomous Driving,'' in Proc. Conf. Language Modeling (COLM), 2024.

\bibitem{ref19} Z. Huang, Z. Sheng, Y. Qu, J. You, and S. Chen, ``VLM-RL: A unified vision language models and reinforcement learning framework for safe autonomous driving,'' \emph{Transportation Research Part C: Emerging Technologies}, vol. 180, art. 105321, 2025, doi: 10.1016/j.trc.2025.105321.

\bibitem{ref20} S. Wang et al., ``OmniDrive: A Holistic Vision-Language Dataset for Autonomous Driving with Counterfactual Reasoning,'' in Proc. IEEE/CVF CVPR, 2025, pp. 22442--22452, doi: 10.1109/CVPR52734.2025.02090.

\bibitem{ref21}
K. A. Mustafa, D. J. Ornia, J. Kober, and J. Alonso-Mora,
``RACP: Risk-aware contingency planning with multi-modal predictions,''
\emph{IEEE Transactions on Intelligent Vehicles}, vol. 10, no. 1,
pp. 228--243, 2025,
doi: 10.1109/TIV.2024.3411530.

\bibitem{ref27} H. Shao et al., ``LMDrive: Closed-Loop End-to-End Driving with Large Language Models,'' in Proc. IEEE/CVF CVPR, 2024, pp. 15120--15130, doi: 10.1109/CVPR52733.2024.01432.

\bibitem{ref28} S. Lai, Z. Xu, W. Zhang, H. Liu, and H. Xiong, ``LLMLight: Large Language Models as Traffic Signal Control Agents,'' in Proc. 31st ACM SIGKDD Conf. Knowledge Discovery and Data Mining, 2025, pp. 2335--2346, doi: 10.1145/3690624.3709379.

\bibitem{ref29} M. Movahedi and J. Choi, ``The Crossroads of LLM and Traffic Control: A Study on Large Language Models in Adaptive Traffic Signal Control,'' IEEE Transactions on Intelligent Transportation Systems, vol. 26, no. 2, pp. 1701--1716, 2025, doi: 10.1109/TITS.2024.3498735.

\bibitem{ref30} S. Masri, H. I. Ashqar, and M. Elhenawy, ``Large Language Models as Traffic Control Systems at Urban Intersections: A New Paradigm,'' Vehicles, vol. 7, art. 11, 2025, doi: 10.3390/vehicles7010011.

\bibitem{ref31} H. Wei et al., ``PressLight: Learning Max Pressure Control to Coordinate Traffic Signals in Arterial Network,'' in Proc. 25th ACM SIGKDD Conf. Knowledge Discovery and Data Mining, 2019, pp. 1290--1298, doi: 10.1145/3292500.3330949.

\bibitem{ref32} X. Wang et al., ``Traffic Light Optimization With Low Penetration Rate Vehicle Trajectory Data,'' Nature Communications, vol. 15, art. 1306, 2024, doi: 10.1038/s41467-024-45427-4.

\bibitem{ref34} M. Abu Tami, H. I. Ashqar, M. Elhenawy, S. Glaser, and A. Rakotonirainy, ``Using Multimodal Large Language Models for Automated Detection of Traffic Safety-Critical Events,'' Vehicles, vol. 6, pp. 1571--1590, 2024, doi: 10.3390/vehicles6030074.

\bibitem{ref35}
Y. Kim, A. S. Abdelrahman, and M. Abdel-Aty,
``VRU-Accident: A vision--language benchmark for video question
answering and dense captioning for accident scene understanding,''
in \emph{Proc. IEEE/CVF ICCV Workshops}, 2025, pp. 772--782,
doi: 10.1109/ICCVW69036.2025.00085.

\bibitem{ref36}
Z. Sheng et al.,
``SafePLUG: Empowering multimodal LLMs with pixel-level insight and
temporal grounding for traffic accident understanding,''
\emph{CHAIN}, vol. 3, no. 1, pp. 53--72, 2026,
doi: 10.23919/CHAIN.2026.000005.

\bibitem{ref37} M. Mumtarin, M. S. Chowdhury, and J. Wood, ``Large Language Models in Analyzing Crash Narratives---A Comparative Study of ChatGPT, BARD and GPT-4,'' arXiv:2308.13563, 2023.

\bibitem{ref38} L. Wang et al., ``Agentic Large Language Models for Day-to-Day Route Choices,'' Transportation Research Part C: Emerging Technologies, vol. 180, art. 105307, 2025, doi: 10.1016/j.trc.2025.105307.

\bibitem{ref39} S. Devunuri and L. J. Lehe, ``TransitGPT: A Generative AI-Based Framework for Interacting with GTFS Data Using Large Language Models,'' Public Transport, vol. 17, no. 2, pp. 319--345, 2025, doi: 10.1007/s12469-025-00395-w.

\bibitem{ref40} S. Li, T. Azfar, and R. Ke, ``ChatSUMO: Large Language Model for Automating Traffic Scenario Generation in Simulation of Urban MObility,'' IEEE Transactions on Intelligent Vehicles, vol. 10, no. 11, pp. 4962--4973, 2025, doi: 10.1109/TIV.2024.3508471.

\bibitem{ref41} M. Jeong, J. Chang, and Y. Yoon, ``Speak to Simulate: An LLM-Guided Agentic Framework for Traffic Simulation in SUMO,'' in Proc. 8th ACM SIGSPATIAL International Workshop on Geospatial Simulation, 2025, pp. 45--48, doi: 10.1145/3764921.3770151.

\bibitem{ref42} X. Guo, X. Yang, M. Peng, H. Lu, M. Zhu, and H. Yang, ``Automating Traffic Model Enhancement With AI Research Agent,'' Transportation Research Part C: Emerging Technologies, vol. 178, art. 105187, 2025, doi: 10.1016/j.trc.2025.105187.

\bibitem{ref43} K. Greshake, S. Abdelnabi, S. Mishra, C. Endres, T. Holz, and M. Fritz, ``Not What You've Signed Up For: Compromising Real-World LLM-Integrated Applications with Indirect Prompt Injection,'' in Proc. ACM Workshop on AI and Security, 2023, pp. 79--90, doi: 10.1145/3605764.3623985.

\bibitem{ref44} ISO, ``Road Vehicles---Functional Safety---Part 1: Vocabulary,'' ISO 26262-1:2018, 2018.

\bibitem{ref45} ISO, ``Road Vehicles---Safety of the Intended Functionality,'' ISO 21448:2022, 2022.

\bibitem{ref46} ISO and SAE International, ``Road Vehicles---Cybersecurity Engineering,'' ISO/SAE 21434:2021, 2021.

\bibitem{ref47} ISO, ``Road Vehicles---Safety and Artificial Intelligence,'' ISO/PAS 8800:2024, 2024.

\bibitem{ref48} IEEE, ``IEEE Standard for Transparency of Autonomous Systems,'' IEEE Std 7001-2021, 2021.

\bibitem{ref49} E. Tabassi, ``Artificial Intelligence Risk Management Framework (AI RMF 1.0),'' NIST AI 100-1, National Institute of Standards and Technology, 2023, doi: 10.6028/NIST.AI.100-1.

\bibitem{ref50} ISO/IEC, ``Information Technology---Artificial Intelligence---Guidance on Risk Management,'' ISO/IEC 23894:2023, 2023.

\bibitem{ref51} ISO/IEC, ``Information Technology---Artificial Intelligence---Management System,'' ISO/IEC 42001:2023, 2023.

\bibitem{ref52}
H. Caesar et al.,
``nuScenes: A multimodal dataset for autonomous driving,''
in \emph{Proc. IEEE/CVF CVPR}, 2020, pp. 11618--11628,
doi: 10.1109/CVPR42600.2020.01164.

\bibitem{ref53}
P. Sun et al.,
``Scalability in perception for autonomous driving: Waymo Open
Dataset,''
in \emph{Proc. IEEE/CVF CVPR}, 2020, pp. 2443--2451,
doi: 10.1109/CVPR42600.2020.00252.

\bibitem{ref54} R. Krajewski et al., ``The highD Dataset: A Drone Dataset of Naturalistic Vehicle Trajectories on German Highways for Validation of Highly Automated Driving Systems,'' in Proc. IEEE ITSC, 2018, pp. 2118--2125, doi: 10.1109/ITSC.2018.8569552.

\bibitem{ref55} H. Zhang et al., ``CityFlow: A Multi-Agent Reinforcement Learning Environment for Large Scale City Traffic Scenario,'' in Proc. WWW, 2019, pp. 3620--3624, doi: 10.1145/3308558.3314139.

\bibitem{ref56} P. A. Lopez et al., ``Microscopic Traffic Simulation Using SUMO,'' in Proc. IEEE ITSC, 2018, pp. 2575--2582, doi: 10.1109/ITSC.2018.8569938.

\bibitem{ref57}
L. Gan, W. Chu, G. Li, X. Tang, and K. Li,
``Large models for intelligent transportation systems and autonomous
vehicles: A survey,''
\emph{Advanced Engineering Informatics}, vol. 62, art. 102786, 2024,
doi: 10.1016/j.aei.2024.102786.

\bibitem{ref58}
C. Zhang, B. Wei, and L. Yang,
``TrafficMind: A system-oriented review of large language models for
intelligent transportation systems,''
\emph{Journal of Transportation Engineering, Part A: Systems},
vol. 152, no. 9, art. 03126005, 2026,
doi: 10.1061/JTEPBS.TEENG-9794.

\bibitem{ref59}
S. Fourati, W. Jaafar, N. Baccar, S. Alfattani, and R. Langar,
``Foundation models for autonomous driving: A comprehensive survey,''
\emph{Engineering Applications of Artificial Intelligence},
vol. 176, art. 114805, 2026,
doi: 10.1016/j.engappai.2026.114805.

\bibitem{ref60}
V. Hassija, T. Majumder, D. Roy, R. Piyush, and V. Chamola,
``The role of large language models (LLMs) in enhancing intelligent
transportation systems: A survey,''
\emph{Vehicular Communications}, vol. 58, art. 100996, 2026,
doi: 10.1016/j.vehcom.2025.100996.

\bibitem{ref61}
D. Mahmud, H. Hajmohamed, S. Almentheri, S. Alqaydi, L. Aldhaheri,
R. A. Khalil, and N. Saeed,
``Integrating LLMs with ITS: Recent advances, potentials, challenges,
and future directions,''
\emph{IEEE Transactions on Intelligent Transportation Systems},
vol. 26, no. 5, pp. 5674--5709, 2025,
doi: 10.1109/TITS.2025.3528116.


\bibitem{ref62}
L. Wen et al., ``DiLu: A Knowledge-Driven Approach to Autonomous Driving
with Large Language Models,'' in \emph{Proc. Int. Conf. Learning
Representations (ICLR)}, 2024.

\bibitem{ref63}
Y. Ma, Y. Cao, J. Sun, M. Pavone, and C. Xiao, ``Dolphins: Multimodal
Language Model for Driving,'' in \emph{Computer Vision--ECCV 2024},
2024, doi: 10.1007/978-3-031-72995-9\_23.

\bibitem{ref64}
B. Li et al., ``Driving Everywhere with Large Language Model Policy
Adaptation,'' in \emph{Proc. IEEE/CVF CVPR}, 2024, pp. 14948--14957.

\bibitem{ref65}
Y. Ma et al., ``LaMPilot: An Open Benchmark Dataset for Autonomous
Driving with Language Model Programs,'' in \emph{Proc. IEEE/CVF CVPR},
2024, pp. 15141--15151.

\bibitem{ref66}
J. Zhang, C. Xu, and B. Li, ``ChatScene: Knowledge-Enabled
Safety-Critical Scenario Generation for Autonomous Vehicles,'' in
\emph{Proc. IEEE/CVF CVPR}, 2024, pp. 15459--15469.

\bibitem{ref67}
Y. Wei et al., ``Editable Scene Simulation for Autonomous Driving via
Collaborative LLM-Agents,'' in \emph{Proc. IEEE/CVF CVPR}, 2024,
pp. 15077--15087.

\bibitem{ref68}
C. Chang, S. Wang, J. Zhang, J. Ge, and L. Li, ``LLMScenario: Large
Language Model Driven Scenario Generation,'' \emph{IEEE Trans. Syst.,
Man, Cybern.: Syst.}, vol. 54, no. 11, pp. 6581--6594, 2024,
doi: 10.1109/TSMC.2024.3392930.

\bibitem{ref69}
X. Tian et al., ``DriveVLM: The Convergence of Autonomous Driving and
Large Vision-Language Models,'' in \emph{Proc. 8th Conf. Robot
Learning}, PMLR, vol. 270, pp. 4698--4726, 2025.

\bibitem{ref70}
Z. Xu et al., ``DriveGPT4-V2: Harnessing Large Language Model
Capabilities for Enhanced Closed-Loop Autonomous Driving,'' in
\emph{Proc. IEEE/CVF CVPR}, 2025, pp. 17261--17270.

\bibitem{ref71}
H. Fu et al., ``ORION: A Holistic End-to-End Autonomous Driving
Framework by Vision-Language Instructed Action Generation,'' in
\emph{Proc. IEEE/CVF ICCV}, 2025, pp. 24823--24834.

\bibitem{ref72}
Q. Liu, C. Li, and W. Ma, ``GATSim: Urban Mobility Simulation with
Generative Agents,'' \emph{Transportation Research Part C: Emerging
Technologies}, vol. 186, art. 105576, 2026,
doi: 10.1016/j.trc.2026.105576.

\bibitem{ref73}
Z. Zhou and S. Zhang, ``AITP: Traffic Accident Responsibility Allocation
via Multimodal Large Language Models,'' in \emph{Proc. IEEE/CVF CVPR
Findings}, 2026, pp. 9259--9268.

\bibitem{ref74}
L. Zhang et al., ``MindDriver: Introducing Progressive Multimodal
Reasoning for Autonomous Driving,'' in \emph{Proc. IEEE/CVF CVPR},
2026, pp. 17831--17841.

\bibitem{ref75}
X. Luo et al., ``V2X-UniPool: Unifying Multimodal Perception and
Knowledge Reasoning for Autonomous Driving,'' in \emph{Proc. IEEE/CVF
CVPR Workshops}, 2026, pp. 747--756.

\bibitem{ref76}
Y. Jin and J. Ma, ``Large Language Model as Parking Planning Agent in
the Context of Mixed Period of Autonomous Vehicles and Human-Driven
Vehicles,'' \emph{Sustainable Cities and Society}, vol. 117,
art. 105940, 2024, doi: 10.1016/j.scs.2024.105940.

\bibitem{ref77}
A. U. Z. Patwary et al., ``Bridging AI and Traffic Simulation: A Robust
and Comprehensive Framework for LLM-Based AI Replanning Agents in
MATSim,'' \emph{Procedia Computer Science}, vol. 280, pp. 622--629,
2026, doi: 10.1016/j.procs.2026.04.079.

\bibitem{ref78}
X. Luo, G. Xu, A. Saroj, J. Yuan, P. Kadav, Y. Shao, and C. R. Wang,
``Agentic Traffic Intelligence: Augmented Human-in-the-Loop Scenario
Generation for Microscopic Traffic Simulation,'' \emph{Artificial
Intelligence for Transportation}, vol. 6, art. 100057, 2026,
doi: 10.1016/j.ait.2026.100057.

\bibitem{ref79}
Y. Ding, M. Maniparambil, N. E. O'Connor, and M. Liu,
``Large Language Model-Assisted Multi-Objective Optimization for an
Integrated Multimodal E-Mobility Platform,'' \emph{Transportation
Research Interdisciplinary Perspectives}, vol. 37, art. 101948, 2026,
doi: 10.1016/j.trip.2026.101948.

\bibitem{ref80}
Q. Wang, S. Wei, and K. Yang,
``LLMs as Virtual Traffic Police: Incident-Aware Traffic Signal Control
Augmented by Large Language Models,'' in \emph{Proc. IEEE 28th Int. Conf.
Intell. Transp. Syst. (ITSC)}, 2025,
doi: 10.1109/ITSC60802.2025.11423524.

\bibitem{ref81}
R. Li et al., ``An Efficient Simulation Scene Generation Method Based on
Extracted Road Network Topology and Large Language Models,''
\emph{Future Transportation}, vol. 6, no. 2, art. 81, 2026,
doi: 10.3390/futuretransp6020081.

\bibitem{ref82}
Z. Xu et al., ``DriveGPT4: Interpretable End-to-End Autonomous Driving
Via Large Language Model,'' \emph{IEEE Robotics and Automation
Letters}, 2024, doi: 10.1109/LRA.2024.3440097.

\bibitem{ref83}
S. Li, M. Ma, T. Azfar, and R. Ke, ``ChatSUMO Agent: An LLM-Based Agent
for Conversational Traffic Simulation in SUMO,'' SSRN preprint,
1 Jan. 2026, doi: 10.2139/ssrn.6000335.

\bibitem{ref84}
Z. Dong, Y. Zhu, Y. Li, K. Mahon, and Y. Sun,
``Generalizing End-to-End Autonomous Driving in Real-World Environments
Using Zero-Shot LLMs,'' in \emph{Proc. 8th Conf. Robot Learning},
PMLR, vol. 270, pp. 1231--1249, 2025. [Online]. Available:
\url{https://proceedings.mlr.press/v270/dong25a.html}

\bibitem{ref85}
M. Cercola, N. Gatti, P. Huertas Leyva, B. Carambia, and S. Formentin,
``Automating the Loop in Traffic Incident Management on Highway,'' in
\emph{Proc. 7th Annu. Learning for Dynamics and Control Conf.}, PMLR,
vol. 283, pp. 272--284, 2025. [Online]. Available:
\url{https://proceedings.mlr.press/v283/cercola25a.html}

\bibitem{ref90}
S. Tan, B. Ivanovic, Y. Chen, B. Li, X. Weng, Y. Cao, P. Kraehenbuehl,
and M. Pavone, ``Promptable Closed-Loop Traffic Simulation,'' in
\emph{Proc. 8th Conf. Robot Learning}, PMLR, vol. 270, pp. 5087--5105,
2025. [Online]. Available:
\url{https://proceedings.mlr.press/v270/tan25a.html}

\bibitem{ref91}
E. Ma et al., ``DriveCombo: Benchmarking Compositional Traffic Rule
Reasoning in Autonomous Driving,'' in \emph{Proc. IEEE/CVF Conf. Comput.
Vis. Pattern Recognit. (CVPR)}, pp. 32113--32123, 2026. [Online]. Available:
\url{https://openaccess.thecvf.com/content/CVPR2026/html/Ma_DriveCombo_Benchmarking_Compositional_Traffic_Rule_Reasoning_in_Autonomous_Driving_CVPR_2026_paper.html}


\end{thebibliography}
\end{document}